\documentclass[lettersize,journal]{IEEEtran}
\usepackage{amsmath,amsfonts}
\usepackage{algorithmic}
\usepackage{algorithm}
\usepackage{array}
\usepackage[caption=false,font=normalsize,labelfont=sf,textfont=sf]{subfig}
\usepackage{textcomp}
\usepackage{stfloats}
\usepackage{url}
\usepackage{verbatim}
\usepackage{graphicx}
\usepackage{cite}

\usepackage{array,tabularx}
\usepackage{multirow}
\usepackage{booktabs}
\usepackage{color}
\definecolor{rv1}{RGB}{0,0,0}
\definecolor{md}{RGB}{0,0,0}
\definecolor{rv2}{RGB}{0,0,0}
\definecolor{fn}{RGB}{0,0,0}
\usepackage{makecell}
\usepackage[colorlinks,
            linkcolor=blue,
            anchorcolor=blue,
            citecolor=blue]{hyperref}

\begin{document}

\title{%
{\small
This work has been accepted for publication in IEEE Transactions on Systems, Man, and Cybernetics: Systems. © 2026 IEEE. Personal use of this material is permitted. Permission from IEEE must be obtained for all other uses. DOI: 10.1109/TSMC.2026.3655553\\[0.8em]
}
A Segmentation-driven Editing Method for Bolt Defect Augmentation and Detection
}

\author{Yangjie Xiao,
\thanks{Research supported by the National Natural Science Foundation of China under grant numbers 62076093, 61871182, 62206095, by the Fundamental Research Funds for the Central Universities under grant numbers 2022MS078, 2023JG002, 2023JC006. \textcolor{fn}{\textit{(Corresponding Author: Ke Zhang)}}}
\thanks{\textcolor{fn}{Yangjie Xiao is with the School of Electrical and Electronic Engineering, North China Electric Power University, Beijing 102206, China.}}
\and{Ke Zhang,~\IEEEmembership{Member,~IEEE,}
\thanks{\textcolor{fn}{Ke Zhang, Yurong Guo, and Zhenbing Zhao are with the Yanzhao Electric Power Laboratory, the Department of Electronic and Communication Engineering, the Hebei Key Laboratory of Power Internet of Things Technology, and the Hebei Engineering Research Center of Intelligent Technology for Power Internet of Things, North China Electric Power University, Baoding 071003, Hebei, China(e-mail: zhangkeit@ncepu.edu.cn).}}}
\and{Jiacun Wang,~\IEEEmembership{Senior,~IEEE,}
\thanks{\textcolor{fn}{Jiacun Wang} is with the Computer Science and Software Engineering Department, Monmouth University, West Long Branch 07764, USA.}}
\and{Xin Sheng,
\thanks{\textcolor{fn}{Xin Sheng, Meijuan Chen, Zehua Ren, and Zhaoye Zheng are with the Department of Electronic and Communication Engineering, North China Electric Power University, Baoding 071003, Hebei, China.}}
}
\and{Yurong Guo,
}

\and{Meijuan Chen,
}
\and{Zehua Ren,
}
\and{Zhaoye Zheng,
}
\and{Zhenbing Zhao,~\IEEEmembership{Member,~IEEE,}
}}

\markboth{Journal of \LaTeX\ Class Files,~Vol.~14, No.~8, August~2021}%
{Shell \MakeLowercase{\textit{et al.}}: A Sample Article Using IEEEtran.cls for IEEE Journals}

\maketitle

\begin{abstract}
Bolt defect detection is critical to ensure the safety of transmission lines. However, the scarcity of defect images and imbalanced data distributions significantly limit detection performance. \textcolor{rv1}{To address this issue, we propose a segmentation-driven bolt defect editing method for defect generation and data augmentation. The proposed framework comprises three main components. First, a bolt attribute segmentation model is constructed, which integrates contrast-based enhancement and frequency-domain filtering to improve edge feature representation. A multipart-aware optimization strategy is introduced to handle the non-contiguous structures of bolt attributes, enabling the generation of high-quality structural masks for subsequent editing. Second, a bolt attribute editing module based on image inpainting is designed, which removes specific bolt attributes by applying mask boundary optimization and context-aware reconstruction, thereby transforming normal bolts into defective ones. Finally, an editing recovery augmentation strategy is proposed to restore the edited bolts into the original inspection images, constructing complete defect samples for expanding the detection dataset. Extensive experiments were conducted on a self-built bolt dataset and the public MVTec AD dataset. The results show that the proposed method significantly outperforms existing state-of-the-art image editing approaches in generation \textcolor{fn}{quality and} effectively improves defect classification and detection performance across multiple mainstream models, thereby validating the effectiveness and practical potential of the proposed approach.} The code of the project is available at https://github.com/Jay-xyj/SBDE. 
\end{abstract}

\begin{IEEEkeywords}
Defect attribute editing, bolt defect generation, SAM, data augmentation, bolt defect detection.
\end{IEEEkeywords}

{\color{rv2}
\section*{Nomenclature}
\addcontentsline{toc}{section}{Nomenclature}
\begin{IEEEdescription}[\IEEEusemathlabelsep\IEEEsetlabelwidth{DFMGAN}] 
    \item[Ada] Adaptive discriminator augmentation.
    \item[DB] DreamBooth.
    \item[E4e] Encoder for editing.
    \item[HS] High-quality SAM.
    \item[NP] Negative-prompt \textcolor{fn}{inversion}.
    \item[NT] Null-text \textcolor{fn}{inversion}.
    \item[PnP] Plug-and-\textcolor{fn}{play inversion}.
    \item[RS] Robust SAM.
    \item[SA] SAM adapter.
    \item[SAM] Segment anything model.
    \item[SBDE] Segmentation-driven bolt defect editing.
    \item[SD] Stable Diffusion.
    \item[SG] StarGAN v2.
\end{IEEEdescription}
}

\section{Introduction}
\IEEEPARstart{B}{olts} are critical components for connecting and securing transmission lines, and their condition directly affects the stability and safety of the power system. Due to long-term exposure to harsh high-altitude environments, bolts are prone to defects such as missing pins or nuts caused by natural aging, environmental corrosion, and external forces. Therefore, the timely and accurate detection of bolt defects is essential to ensure the safe operation of transmission lines \cite{ref-ubddm}. Traditional manual inspection methods are inefficient and difficult to meet the needs of large-scale transmission lines. With the rapid advancement of drone intelligent inspection technologies, defect detection based on deep learning has become a primary approach in transmission line operations and maintenance \cite{ref-dete,ref-zhou,ref-dist}. However, the number of defect images in current research and practical engineering is severely limited, restricting the performance of data-driven detection algorithms. Therefore, effectively expanding defect datasets is an urgent research topic.

In recent years, some studies have focused on expanding defect datasets through generative models \cite{ref-indu}. These models fit the distribution of the original data and generate defect images directly by sampling from noise. SDGAN \cite{ref-sdgan} proposed a surface defect image generation method based on generative adversarial networks (GANs). It generates high-quality and diverse defect images from industrial defect-free images for expanding defect datasets. \textcolor{rv2}{However, such methods usually rely on a sufficient number of defect samples for training, and the generated results are difficult to control in terms of defect location. Moreover, when the number of defect samples is limited, the generation process is easily susceptible to random noise, leading to structural distortion, which makes it difficult to satisfy the requirements for defect morphology and structural consistency in complex industrial scenarios.}

It is worth noting that, although defect images are scarce, normal images are sufficient and share a high degree of feature correlation with defect images. Therefore, generating defect images through attribute editing of normal images is a more efficient and highly promising solution. However, the implementation of such methods faces challenges in capturing few-shot defect features. Currently, the conversion of normal images into defect images has become a research hotspot in the industrial field. DFMGAN \cite{ref-dfmgan} proposed a defect image generation method for few-sample scenarios. By training a data-efficient generative model on defect-free images and incorporating a defect-aware residual module, it achieves high-quality and diverse defect image generation. Although these methods have achieved good results, they are usually applicable to images with simple structures or uniform backgrounds. They focus on transforming one class of normal images into corresponding defect images rather than performing precise attribute editing on the same normal image. Even when defects are added through generative sampling and the original image is reconstructed, the original image is inevitably changed \cite{ref-e4e}.

More importantly, these existing defect conversion methods are not suited for bolt defect scenarios in the power field. \textcolor{rv2}{The fundamental reason lies in the structural and quality limitations of bolt images. On the one hand,} bolts exhibit highly regular geometric shapes and defect types, necessitating editing methods that achieve high-precision attribute conversion while meeting industrial standards \cite{ref-avsc}. \textcolor{rv2}{On the other hand,} bolt images are typically small, with minimal pixel contrast between the bolt and its background, significantly increasing the difficulty for generative models to edit the target regions precisely. \textcolor{rv2}{Based on the above characteristics, bolt defect conversion differs from industrial surface defect generation, and even slight deviations can cause structural distortions,} severely affecting the editing quality. Therefore, achieving accurate attribute editing for bolts requires overcoming many challenges, including precise localization of target regions, maintaining background consistency, and ensuring accurate \textcolor{fn}{edits} of defect attributes.

\begin{figure}[!t]
\centering
\includegraphics[width=1.0\linewidth]{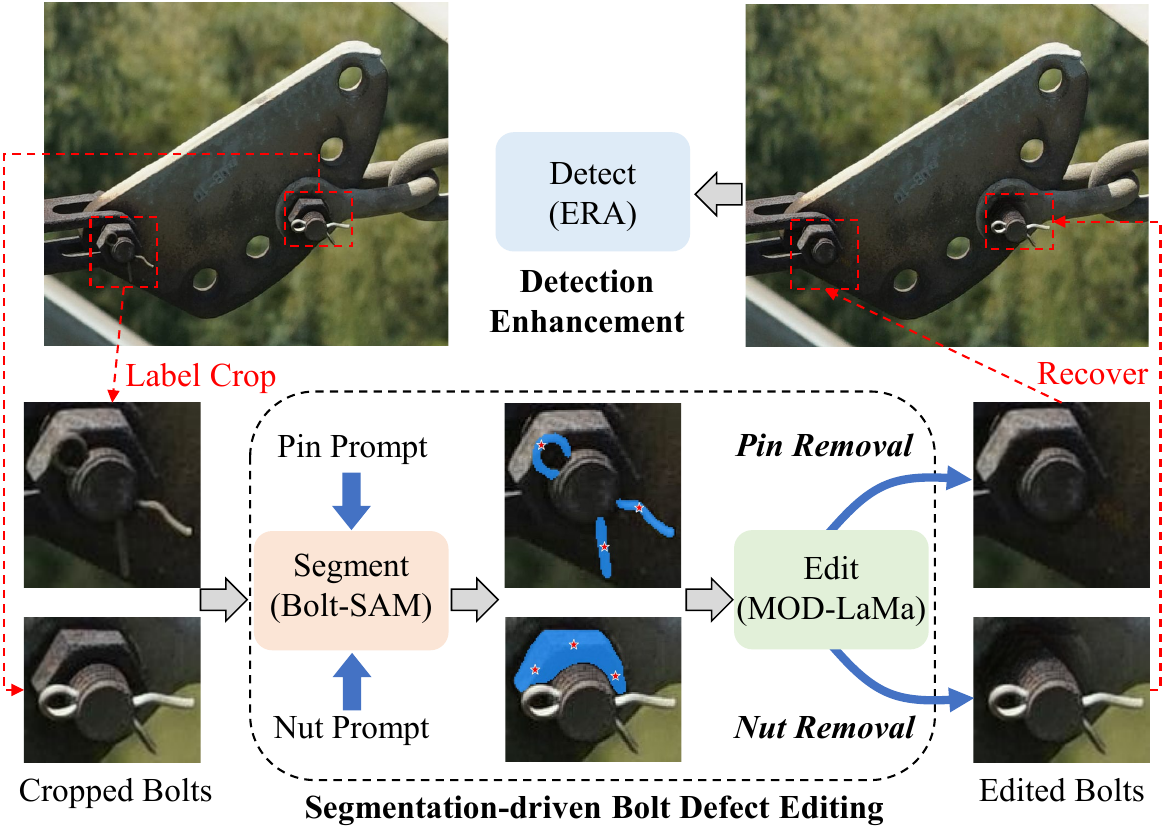}
\vspace{-0.6cm}
\caption{\textcolor{rv2}{The overall framework of the proposed method.}}
\label{fig1}
\vspace{-0.3cm}
\end{figure}

To address the above issues, we propose a segmentation-driven bolt defect editing method (SBDE), as illustrated in Fig. \ref{fig1}. \textcolor{rv2}{The method leverages the structural characteristics of transmission line bolts to guide defect editing by extracting their key attribute regions. It precisely transforms normal bolts into defective ones and embeds the results into the original inspection images to construct an augmented dataset that supports downstream tasks.} The contributions of this paper are summarized as follows:

\begin{enumerate}{}{}
\item{\textcolor{rv2}{We construct the first attribute-level defect editing framework for transmission line bolts. It connects attribute segmentation, defect editing, and data augmentation into a cross-task joint \textcolor{fn}{closed-loop}, providing a solution to defect sample scarcity in power vision.}}
\item{\textcolor{rv2}{We propose a bolt attribute segmentation model, Bolt-SAM, which integrates contrast-limited adaptive histogram equalization and frequency-domain features to enhance edge details, and designs a multipart-aware mask decoder to accurately segment non-contiguous bolt attributes.}}
\item{\textcolor{rv2}{We design a bolt defect editing module, MOD-LaMa, which enhances contextual information through mask boundary optimization and achieves defect editing by removing specified bolt attributes via image inpainting.}}
\item{\textcolor{rv2}{We propose an editing recovery augmentation strategy, ERA, which restores edited defect bolts into original inspection images to construct realistic defect samples, effectively expanding the dataset and significantly improving the performance of mainstream detection and classification models.}}
\end{enumerate}

The rest of this paper is organized as follows. We first review the related work in Section II. Then, Section III describes our method in detail. After that, Section IV introduces our experiment and the experimental results. Finally, we summarize the work and discuss future work in Section V.

\section{Related Works}
\subsection{Segment Anything Model}
Segment Anything Model (SAM) \cite{ref-sam} is a powerful foundational segmentation model that supports multi-modal inputs, including point prompts, box prompts, and mask prompts. It can perform zero-shot segmentation in unknown domains and demonstrates excellent performance in various vision tasks. High-quality SAM (HS) \cite{ref-hqsam} introduced high-quality output tokens by fusing early and final ViT features, improving the segmentation performance of SAM while maintaining its efficiency and zero-shot generalizability. SAM-Adapter (SA) \cite{ref-samadapter} proposed a simple and efficient adapter mechanism that integrates domain-specific information with visual prompts, enhancing SAM's performance in complex segmentation tasks without requiring fine-tuning of the SAM network. For low-quality image scenes, RobustSAM (RS) \cite{ref-rbsam} proposed a series of anti-degradation modules to enhance the segmentation robustness of SAM under various image degradation conditions.

Moreover, SAM’s powerful capabilities have been extended to other tasks. Inpaint Anything (IA) \cite{ref-ia} proposed a new ``click-and-fill'' paradigm that combines SAM with various models to achieve precise localization and inpainting of target regions. \textcolor{rv2}{In specific domains, several studies have applied SAM or its variants to medical and industrial segmentation tasks. Ma et al. \cite{ref-medsam} retrained SAM on medical images, leading to significant improvement in \textcolor{fn}{organ and lesion} segmentation accuracy. Teng et al. \cite{ref-bcsam} introduced a prompting method based on a fractal dimension matrix for automatic segmentation of concrete bridge cracks. Nonetheless, the power domain still lacks fine-grained segmentation research focusing on equipment component attributes. How to integrate general segmentation models with domain-specific structural priors to achieve attribute-level localization and structure-aware segmentation of key components, such as bolts, remains an important topic for further research.}

\subsection{Image Editing}
\textcolor{md}{Image editing has long been a key \textcolor{fn}{task} in computer vision. As a classic generative model, \textcolor{fn}{the} Generative Adversarial Network (GAN) has found extensive applications in image editing tasks. StarGANv2 (SG) \cite{ref-stargan2} proposed multi-domain translation frameworks, supporting the transformation of multiple target attributes. Adaptive discriminator augmentation (Ada) \cite{ref-ada} set a benchmark in image editing by incorporating style encoding and generator architecture, allowing the generation of high-quality images with semantic disentanglement. Additionally, Encoder for editing (E4e) \cite{ref-e4e} achieved accurate inversion and flexible editing of real images through latent space optimization.}

In recent years, the rise of diffusion models has further advanced generative technologies. Stable Diffusion (SD) \cite{ref-sd}, as an open-source diffusion model, has been extensively used for generation and editing tasks due to its efficient latent space operations, further driving research in diffusion models. On this basis, various fine-tuning and inversion methods have been proposed to meet the needs of different tasks. DreamBooth (DB) \cite{ref-db} generated high-fidelity images from a small set of reference images while preserving subject features. Regarding diffusion inversion methods, Null-text Inversion (NT) \cite{ref-null} combined pivot inversion with text-free optimization to support flexible text-driven image editing. Negative-prompt Inversion (NP) \cite{ref-nega} achieved efficient editing through forward propagation without requiring complex optimization, while Plug-and-Play Inversion (PnP) \cite{ref-pnp} employed a branching optimization strategy that significantly accelerated inference while improving editing precision. \textcolor{rv2}{Flow-based generative models are also emerging in image editing \cite{ref-flux}. Rectified Flow Editing (RF-Edit) \cite{ref-rfedit} improves rectified flow inversion and ordinary differential equation solving to preserve structure in local edits. FlowEdit \cite{ref-flowedit} proposes a text-driven framework that avoids inversion and optimization, enabling effective editing across multiple pre-trained flow models.}

\textcolor{md}{Given the superior performance of existing generative models, researchers have applied them to defect-generation tasks. DFMGAN \cite{ref-dfmgan} combines a defect-aware residual module to generate diverse defect images in low-sample scenarios. AnoDiff \cite{ref-anodiff} utilized a diffusion model with Spatial Anomaly Embedding and Adaptive Attention Re-weighting to generate high-quality anomaly images in few-shot scenarios.}

Although the above methods perform well in industrial defects, they are not suitable for bolt defect editing. Unlike general surface defects, bolt defects typically manifest as the loss of key attributes, requiring editing methods to precisely remove target regions while maintaining background consistency and preserving the original structural features. Faced with the professionalism and complexity of bolt, attribute editing based on generative models is a great challenge. In contrast, image inpainting-based approaches offer a feasible alternative. \textcolor{rv2}{Large Mask Inpainting (LaMa) \cite{ref-lama} achieves high-quality restoration via fast fourier convolutions and large-mask training, and IA \cite{ref-ia} integrates SAM with LaMa for target region restoration. However, these methods focus on region-level inpainting in natural images and lack structural and semantic constraints for industrial components. Therefore, for the first time, we introduce image inpainting into bolt attribute-level editing to precisely remove component attributes and enable defect conversion in industrial applications.}
\vspace{-0.2cm}
\subsection{Bolt Defect Detection}
Bolt defect detection is one of the key technologies to ensure the safe operation of transmission lines. In recent years, deep learning-based object detection methods have been widely applied in this field. Zhao et al. \cite{ref-avsc} proposed an automatic visual shape clustering network to tackle the complexity of bolt shapes in transmission lines. By leveraging unsupervised clustering alongside feature enhancement, fusion, and regional feature extraction, it significantly improved the detection of missing pins. Luo et al. \cite{ref-ubddm} addressed ultra-small target detection by designing an ultra-small target perception module and a multi-scale feature fusion strategy. Its two-stage detection method enabled accurate bolt defect detection with minimal annotation cost.

\textcolor{md}{Most existing studies focus on architectural optimization, while data-level exploration remains limited, leading to degraded detection performance under insufficient or imbalanced defect samples. To address this, we propose the ERA strategy to augment bolt defect data through image editing, effectively alleviating the sample scarcity bottleneck. The method is compatible with various detection and classification models and improves downstream performance.}

\section{Method}
To address the severe imbalance between normal and defective bolt samples in transmission lines, particularly the extreme scarcity of defect samples, we propose a segmentation-driven bolt defect editing method named SBDE. \textcolor{rv2}{The overall framework of the proposed method is illustrated in Fig. \ref{fig1}. Specifically, Bolt-SAM extracts attribute masks of key bolt components to provide structural priors for the editing stage; MOD-LaMa removes specific attributes based on these masks to achieve structured bolt defect editing; and the ERA strategy embeds the edited results back into the original inspection images to expand the dataset and enhance detection performance. SBDE establishes a cross-task joint framework integrating segmentation, editing, and augmentation, enabling the transformation from normal to defective bolts while improving downstream task performance and maintaining} strong compatibility with various models.
\vspace{-0.2cm}
\subsection{Segment: Bolt-SAM}
\begin{figure*}[!t]
\centering
\includegraphics[width=0.8\linewidth]{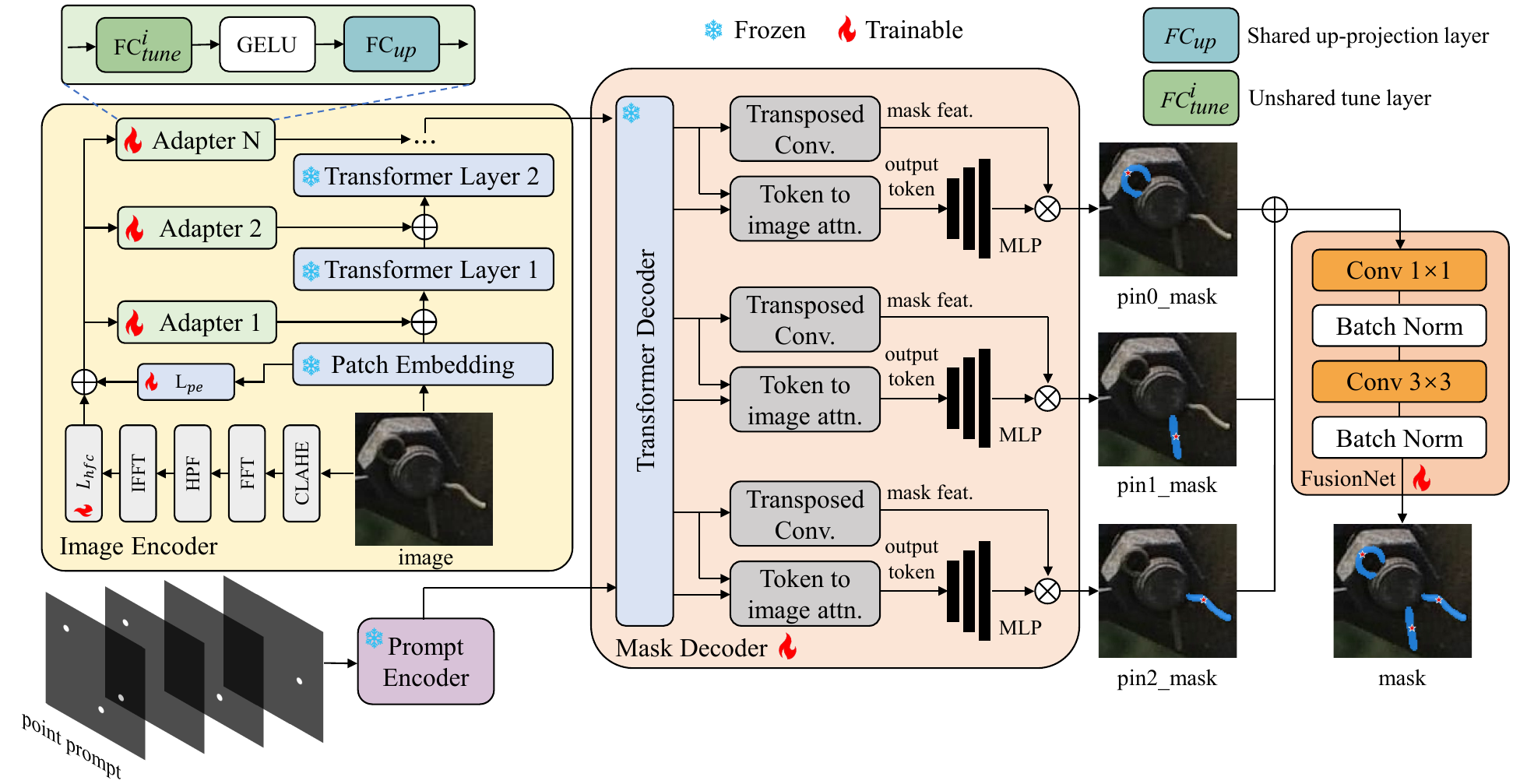}
\vspace{-0.5cm}
\caption{\textcolor{rv1}{The architecture of Bolt-SAM. We incorporate the adapter module into each transformer layer to enhance features and adopt a multi-task fine-tuning mask decoder architecture to improve attribute segmentation accuracy. \textcolor{fn}{FusionNet, built with a simple MLP,} integrates the complete attribute mask. The parameters of the original image encoder and prompt encoder are frozen, while the adapter module, mask decoder, and fusion network are set trainable.}}
\label{fig2}
\vspace{-0.3cm}
\end{figure*}

In the SBDE model, the goal of the segmentation stage is to accurately extract key attribute masks from normal bolt images, providing precise segmentation regions for subsequent defect editing. To achieve this, we propose an improved segmentation model called Bolt-SAM, which is based on the baseline RobustSAM. Its structure is illustrated in Fig. \ref{fig2}. \textcolor{rv1}{It consists of two main modules: the CLAHE-FFT Adapter (CFA) and the Multipart-Aware Mask Decoder (MAMD).} \textcolor{rv2}{The CFA is designed to address the common low-contrast and uneven illumination issues in bolt images by introducing contrast-limited adaptive histogram equalization and frequency-domain enhancement to strengthen edge features. The MAMD focuses on non-contiguous bolt attributes such as pins and adopts a multi-part mask generation and fusion strategy to improve overall structural segmentation accuracy.}

Taking the pin attribute of the bolt as an example, the input image first passes through the CFA module, where adaptive frequency domain information is introduced into the image encoder, resulting in enhanced image features. Subsequently, in the MAMD module, the mask decoder, the pin is divided into three parts, and multi-task fine-tuning is performed based on corresponding point prompts to generate local masks for each part. Finally, \textcolor{rv2}{FusionNet} integrates the local masks from each part to generate a complete pin mask, thereby achieving accurate segmentation of the key attributes of the bolt. \textcolor{rv2}{Herein, FusionNet is a lightweight convolutional fusion network with a simple structure, consisting of sequentially connected $1 \times 1$ and $3 \times 3$ convolutional layers followed by \textcolor{fn}{batch normalization}, which effectively ensures boundary and structural consistency in the overall mask.}

\subsubsection{CLAHE-FFT Adapter}
The complex real-world conditions, such as lighting variations, background noise, and low-contrast regions, weaken the model's ability to capture bolt edges, leading to inaccurate segmentation of key attributes in bolt images. To address this issue, we design the CFA module based on the explicit visual prompting approach \cite{ref-evp}, as shown on the left side of the image encoder in Fig. \ref{fig2}. \textcolor{rv2}{In the CFA module, CLAHE is employed to improve lighting and contrast conditions, making the details in low-contrast regions clearer, while FFT preserves low-level visual structures such as edges and contours in the frequency domain to provide additional cues for the encoder. In implementation,} the Adapter is a lightweight multilayer perceptron (MLP), which extracts structural information from both image embeddings and high-frequency components to guide the Transformer’s feature learning, thereby enhancing the representation capability of SAM’s image encoder for bolt attributes.

Specifically, given an input image $I\in \mathbb{R}^{H\times W \times C}$, CFA first enhances details and edge features through Contrast Limited Adaptive Histogram Equalization (CLAHE), resulting in an optimized image $I_{cla}$. $I_{cla}$ is then transformed into the frequency domain by Fast Fourier Transform (FFT) and represented as $z=\text{FFT}(I_{cla})$. CFA constructs a high-pass filter (HPF) using a high-frequency mask $M_{hpf}$ to extract high-frequency edge information. Following the design of the high-frequency mask in reference\cite{ref-evp}, $M_{hpf}$ is defined as:
\begin{equation}
\begin{aligned}
M_{hpf}^{i, j}(\tau)=\left\{\begin{array}{ll}
0, & \frac{4\left|\left(i-\frac{H}{2}\right)\left(j-\frac{W}{2}\right)\right|}{H W} \leq \tau \\
1, & \text { otherwise }
\end{array}\right.
\end{aligned}
\label{eq1}
\end{equation}
\noindent where $H$ and $W$ represent the height and width of the image, respectively, and $\tau$ controls the proportion of the low-frequency region. The high-frequency information is then transformed back to the spatial domain using the Inverse Fast Fourier Transform (IFFT), yielding the corresponding high-frequency component $I_{hfc}=\text{IFFT}(z\cdot M_{hpf})$.

\textcolor{rv1}{Subsequently, the high-frequency component $I_{hfc}$ is fused with the embedding feature $E_{pe}$. $I_{hfc}$ contains structural information such as edges and contours, while the latter encodes the contextual semantics of the image. Since $I_{hfc}$ and $E_{pe}$ have different dimensions, linear projection layers $L_{hfc}$ and $L_{pe}$ project them into a unified feature space. The projected features are then summed to form the input to the adapter:}
\begin{equation}
\begin{aligned}
\textcolor{rv1}{F_i=\mathrm{L}_{hfc}(\mathrm{I}_{hfc})+\mathrm{L}_{pe}(\mathrm{E}_{pe})}
\end{aligned}
\label{eq2}
\end{equation}

\textcolor{rv1}{The Adapter consists of two fully connected (FC) layers and a GELU activation function, designed to learn visual prompts. For the $i$-th adapter, the visual prompt $P_i$ is as follows:}
\begin{equation}
\begin{aligned}
\textcolor{rv1}{P_i=\mathrm{FC}_{up}(\mathrm{GELU}(\mathrm{FC}^i_{tune}(F_i)))}
\end{aligned}
\label{eq3}
\end{equation}
\noindent \textcolor{rv1}{where ${FC}^i_{tune}$ is a tunable layer independently defined for each adapter, which extracts the prompt vector corresponding to the $i$-th Transformer layer from the input features. $FC_up$ is a shared up-projection layer across all adapters, used to match the dimension of the Transformer features. All FC layers are single-layer fully connected networks. Finally, the resulting visual prompt $P_i$ is added to the input feature $X_i$ of the $i$-th Transformer layer to enhance its output feature $Y_i$. This process can be described as follows:}
\begin{equation}
\begin{aligned}
\textcolor{rv1}{X_i=Y_{i-1}+P_i}
\end{aligned}
\label{eq4}
\end{equation}
\begin{equation}
\begin{aligned}
\textcolor{rv1}{Y_i={Transformer}_i(X_i)}
\end{aligned}
\label{eq5}
\end{equation}
\subsubsection{Multipart-Aware Mask Decoder}
The structural and visual complexity of bolt attributes increases the difficulty of SAM segmentation, especially as pins often appear in dispersed forms within images. To address this issue, we propose a segmented fine-tuning architecture called MAMD, illustrated \textcolor{fn}{in} the middle side of Fig. \ref{fig2}. MAMD employs a multi-task fine-tuning strategy, dividing the pin into three parts (pin0, pin1, \textcolor{fn}{and} pin2) for independent optimization, and generates the complete pin mask through a fusion network.

SAM's \textcolor{fn}{box prompts are} not suitable for attribute segmentation in transmission line bolt images. Due to the low resolution of bolt images and the lack of significant contrast between pixels, using \textcolor{fn}{box prompts} tends to introduce excessive information from non-target regions, interfering with segmentation. In contrast, \textcolor{fn}{point prompts provide} more precise guidance, enabling the model to focus on the target region. It is particularly advantageous for non-contiguous structures. Therefore, we fine-tune Bolt-SAM based on \textcolor{fn}{point} prompts. By providing \textcolor{fn}{point prompts} for each local area of the bolt pin, we guide the model to optimize regional features to accurately capture boundary information and detail features, thereby improving the processing accuracy of non-contiguous complex structures. MAMD introduces Focal Loss and Dice Loss as loss functions. For the $i$-th sub-region, the local loss function is defined as:
\begin{equation}
\begin{aligned}
L_{local,i}=\alpha \cdot L_{focal}(M_{i},GT_{i})+\beta \cdot L_{Dice}(M_{i},GT_{i})
\end{aligned}
\label{eq6}
\end{equation}
\noindent where $M_i$ and $GT_i$ represent the predicted mask and ground truth mask for the $i$-th sub-region, respectively, while $\alpha$ and $\beta$ are the weighting coefficients of the loss functions. MAMD integrates local mask features into a complete pin mask using a lightweight fusion network with two convolutional layers and \textcolor{fn}{batch normalization} layers. The optimization of the global mask is the same as the local loss function. \textcolor{rv2}{With a multi-task fine-tuning mechanism, this method} effectively solves the segmentation problem of the non-contiguous structure of bolt pins through multi-task fine-tuning, providing high-quality segmentation results for subsequent bolt defect editing.

\textcolor{rv2}{It is worth noting that the nut structure is relatively complete and continuous, without the discreteness observed in the pin. Therefore, in this case, it is unnecessary to adopt the multi-branch design of MAMD. Instead, a single-branch decoder with point prompts is directly employed to extract the nut mask. This approach maintains segmentation accuracy while avoiding unnecessary complexity. Consistent with the pin segmentation, Focal Loss and Dice Loss are also used to optimize the nut mask.}

\begin{figure}[!t]
\centering
\includegraphics[width=1.0\linewidth]{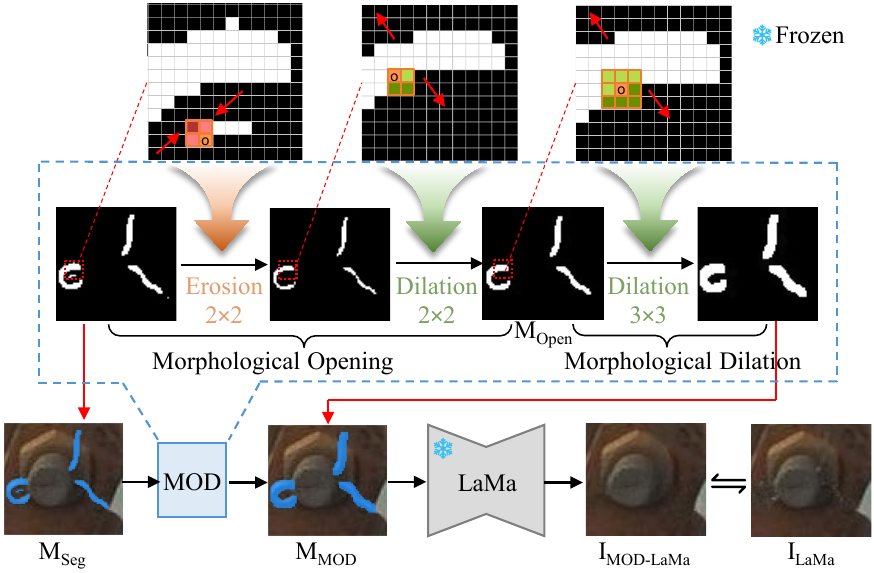}
\vspace{-0.6cm}
\caption{The structure of MOD-LaMa.}
\label{fig3}
\vspace{-0.3cm}
\end{figure}

\subsection{Edit: MOD-LaMa}
Bolt attribute editing addresses the issue of insufficient defect samples by converting normal bolts into defective ones with missing pins or nuts. \textcolor{rv2}{To address this specific type of defect, this study innovatively introduces an image inpainting mechanism into the bolt editing task, achieving defect editing by removing designated bolt attributes. LaMa, which integrates local features with global contextual information, can} generate background-consistent content \textcolor{rv2}{and is thus well suited for the structural restoration of bolt defects.} Due to the low resolution of bolt images, \textcolor{rv2}{which results in insufficient mask coverage or unsmoothed boundaries,} \textcolor{fn}{the effectiveness of LaMa’s inpainting is limited.} We propose MOD-LaMa, as shown in Fig. \ref{fig3}, in which the Morphological Opening and Dilation (MOD) module \textcolor{rv2}{is employed for mask optimization. This module first removes boundary noise through opening operations and then applies dilation to smooth contours and expand target regions, thereby enhancing contextual awareness in local areas.}

\textcolor{md}{We first define the erosion and dilation operations. Let $M$ denote the original mask, $s$ be the structuring element, $(x, y)$ be the coordinates of the origin of $s$, and $(i, j)$ be the offset relative to the origin. The erosion operation $\ominus$ is used to shrink the boundaries of the target region and remove noise within the mask, and is defined as follows:}
\begingroup
\color{rv1}
\vspace{-0.1cm}
\begin{equation}
\begin{aligned}
\textcolor{md}{(M\ominus s)(x,y)=\min_{(x,j)\in s} {M(x+i,y+j)}}
\end{aligned}
\label{eq7}
\end{equation}
\endgroup
\textcolor{rv1}{The dilation operation $\oplus$ restores the integrity of the target region by expanding its boundaries and is defined as follows:}
\begingroup
\color{rv1}
\vspace{-0.1cm}
\begin{equation}
\begin{aligned}
\textcolor{md}{(M\oplus s)(x,y)=\max_{(x,j)\in s} {M(x+i,y+j)}}
\end{aligned}
\label{eq8}
\end{equation}
\endgroup
\textcolor{rv1}{MOD first performs a morphological opening operation, that is, erosion followed by dilation, on the initial segmentation mask $M_{seg}$ to obtain the smoothed mask $M_{open}$:}
\begin{equation}
\begin{aligned}
M_{open}=(M_{seg}\ominus s) \oplus s
\end{aligned}
\label{eq9}
\end{equation}
\textcolor{rv1}{To further expand the contextual region, a second dilation is applied to $M_{open}$ to obtain the final optimized mask $M_{mod}$:}
\begin{equation}
\begin{aligned}
M_{MOD}=M_{open} \oplus s
\end{aligned}
\label{eq10}
\end{equation}
\textcolor{rv1}{Finally, the optimized mask $M_{mod}$ and the original image $I_{ori}$ are fed into LaMa to perform attribute editing for bolt defects, yielding the final edited image $I_{edit}$:}
\begin{equation}
\begin{aligned}
I_{edit}=\mathrm{LaMa}(I_{ori},M_{MOD})
\end{aligned}
\label{eq11}
\end{equation}

\textcolor{md}{MOD-LaMa is critical for bolt defect editing in SBDE, and its efficiency directly impacts deployability. To ensure consistent automated processing, all bolt images are region-aligned and resized to $128 \times 128$. Due to the standardized geometry of transmission line bolts, the preprocessed images maintain consistent spatial scales, enabling the use of fixed structuring elements for batch mask optimization. By default, the opening operation uses a structuring element of size 2, while dilation uses sizes 3 and 5 for pin and nut regions, respectively, which is validated in our experiments.}

\subsection{Detect: ERA}
\begin{figure}[!t]
\centering
\includegraphics[width=0.8\linewidth]{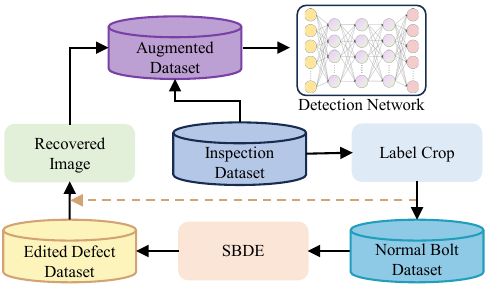}
\caption{The workflow of ERA, which includes inspection image cropping, bolt editing, and defect recovery.}
\label{fig4}
\vspace{-0.3cm}
\end{figure}

\textcolor{rv2}{ To further validate the effectiveness of SBDE, this paper proposes the ERA strategy. By editing defect samples and restoring them to the original inspection scenes, ERA expands the bolt defect detection dataset. This strategy establishes a closed-loop verification mechanism between the editing results and downstream tasks, and the overall process is illustrated in Fig. \ref{fig4}.} Specifically, we use LabelImg to annotate normal bolts in inspection images $I_{ins}$, crop them to extract individual normal bolt $I_{ori}$, and edit them with SBDE to generate defect bolt $I_{edit}$. Finally, $I_{edit}$ is restored to its original position in $I_{ins}$ with the label updated to defect, forming defect-augmented inspection images $I_{aug}$. \textcolor{rv2}{The recovery operation is defined as follows:}
\begin{equation}
\begin{aligned}
I_{aug}(i, j)=\left\{\begin{array}{ll}
I_{edit}(x{'},y{'}), & \text {if} (x,y) \cap {R_{box}} \\
I_{ins}(x,y), & \text { otherwise }
\end{array}\right.
\end{aligned}
\label{eq12}
\end{equation}
\noindent where $R_{box}$ is the bounding box, and $(x{'},y{'})$ refers to the corresponding coordinate mapping in $I_{edit}$.

The ERA-augmented inspection image fully retains the original transmission line scene information, generating defects only in localized bolt regions. This allows the detection model to learn diverse defect features within the same background, enriching the data distribution for defect detection and thereby improving detection performance. Furthermore, ERA avoids over-reliance on real defect data during data augmentation, providing an effective solution for bolt defect detection \textcolor{rv2}{in scenarios with few-shot defect samples.}

\section{Experiments}
\subsection{Datasets}
\begin{figure*}[!t]
\centering
\includegraphics[width=0.8\linewidth]{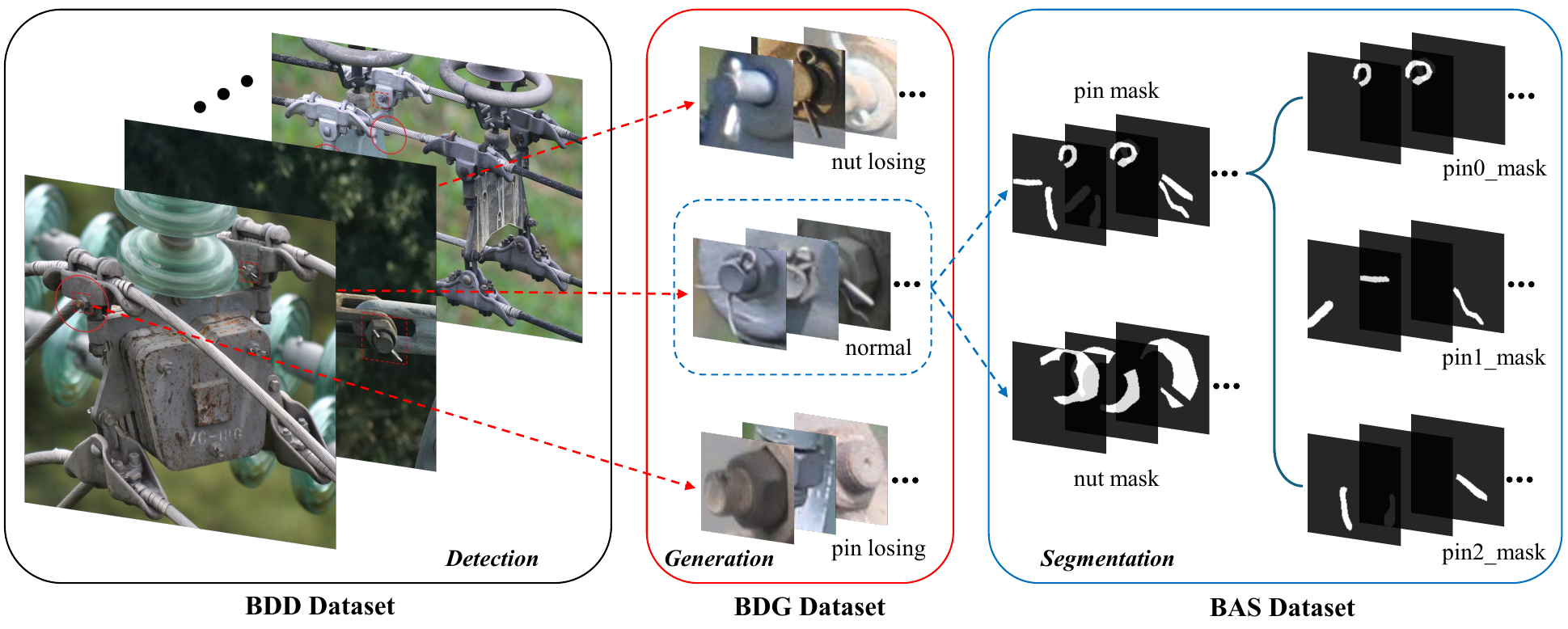}
\vspace{-0.3cm}
\caption{The construction process of the three datasets. They are obtained by annotating, cropping, and filtering the transmission line inspection images, where BAS only contains normal bolts and the pin attributes are divided into multiple masks.}
\label{fig5}
\vspace{-0.3cm}
\end{figure*}

\textcolor{md}{We constructed three datasets for bolt defect detection, generation, and attribute segmentation. All datasets originate from drone-captured transmission line images collected under diverse real-world conditions, ensuring data authenticity and diversity. Their relationships, representative samples, and construction pipeline are shown in Fig. \ref{fig5}.}

The Bolt Defect Detection dataset (BDD) was created by annotating bolts in inspection images using LabelImg, covering three categories: normal (NR), pin losing (PL), and nut losing (NL). As shown in Table \ref{tab1BDD}, it comprises 1770 images with 5835 labeled instances, and its imbalance reflects real-world defect rarity. The Bolt Defect Generation dataset (BDG) was constructed by cropping bolt regions from BDD and filtering out low-quality samples, while reducing normal samples to mitigate class imbalance. As detailed in Table \ref{tab1BDG}, BDG includes 2210 cropped images and is used to train generation methods. To enable attribute segmentation, we further annotated normal samples from BDG using \textcolor{fn}{LabelMe} to build the Bolt Attribute Segmentation dataset (BAS), where the pin region is divided into head (pin0), left pin (pin1), and right pin (pin2). As reported in Table \ref{tab1BAS}, each attribute class includes an equal number of images and masks for segmentation model fine-tuning.

\renewcommand\arraystretch{1}
\begin{table}[!t]
\centering
\caption{Data Distribution of the BDD Datasets.}
\label{tab1BDD}
\begin{tabularx}{\linewidth}{>{\centering\arraybackslash}X>{\centering\arraybackslash}X>{\centering\arraybackslash}X>{\centering\arraybackslash}X>{\centering\arraybackslash}X>{\centering\arraybackslash}X}
\toprule
\multirow{2}{*}{Data} & \multirow{2}{*}{\makecell{Inspection \\images}} & \multicolumn{4}{c}{Instance images} \\ \cmidrule{3-6} 
                      &                                    & NR     & PL    & NL   & All   \\ \midrule
Train                 & 1433                               & 3961   & 449   & 244  & 4654  \\
Test                  & 337                                & 1018   & 100   & 63   & 1181  \\ \bottomrule
\end{tabularx}
\end{table}

\begin{table}[!t]
\centering
\caption{Data Distribution of the BDG Datasets.}
\label{tab1BDG}
\begin{tabularx}{\linewidth}{>{\centering\arraybackslash}X>{\centering\arraybackslash}X>{\centering\arraybackslash}X>{\centering\arraybackslash}X>{\centering\arraybackslash}X}
\toprule
Class & NR   & PL  & NL  & All  \\ \midrule
Train & 1114 & 414 & 240 & 1768 \\
Test  & 279  & 103 & 60  & 442  \\ \bottomrule
\end{tabularx}
\end{table}

\begin{table}[!t]
\centering
\caption{Data Distribution of the BAS Datasets.}
\label{tab1BAS}
\begin{tabularx}{\linewidth}{>{\centering\arraybackslash}X>{\centering\arraybackslash}X>{\centering\arraybackslash}X>{\centering\arraybackslash}X>{\centering\arraybackslash}X>{\centering\arraybackslash}X}
\toprule
Mask  & Nut & Pin & Pin0 & Pin1 & Pin2 \\ \midrule
Train & 886 & 886 & 886  & 886  & 886  \\
Test  & 220 & 220 & 220  & 220  & 220  \\ \bottomrule
\end{tabularx}
\vspace{-0.3cm}
\end{table}

\subsection{Experimental Settings}
During the training, Bolt-SAM fine-tunes only the parameters of CFA and MAMD while freezing the rest of the parameters. Bolt-SAM adopts the points as prompts, randomly sampling three points from the ground truth mask for training. The training configuration includes a learning rate of 0.0001, a batch size of 2, and a total of 50 epochs. The experiments are conducted on two NVIDIA 4090 GPUs, implemented using PyTorch 2.4, CUDA 12.1, and Python 3.10.

\subsection{Evaluation Metrics}\label{sc-metric}
To comprehensively evaluate the performance of SBDE, we selected different evaluation metrics for various tasks. For segmentation, the mean Intersection over Union (mIoU), Dice coefficient (Dice), and Pixel Accuracy (PA) were used to assess the quality of the bolt attribute masks. \textcolor{rv1}{For detection, the mean Average Precision (mAP) across all categories was used to evaluate defect detection performance after applying the ERA strategy. For classification, \textcolor{fn}{accuracy (Acc)},macF1, weightF1, and relative improvement (IMP) were adopted as evaluation metrics to comprehensively reflect the effectiveness of data augmentation using SBDE-edited samples.}

Due to the significant differences between SBDE and traditional generative models in editing, different evaluation metrics are employed in comparative and ablation experiments. For the comparative experiments, we focus on evaluating the quality of the edited images and their overall similarity to the original images. Peak Signal-to-Noise Ratio (PSNR), Structural Similarity Index Measure (SSIM), and Learned Perceptual Image Patch Similarity (LPIPS) are used to measure the differences between the edited images and the original images. \textcolor{rv1}{In addition, Kernel Inception Distance (KID) is adopted to comprehensively evaluate the quality and diversity of the generated images.}

\textcolor{rv1}{In the ablation experiments, editing is applied only to local attribute regions to evaluate the contribution of each module to defect attribute editing. We adopt Class-Conditional Frechet Inception Distance (CSFID) to assess the quality of edited images and their alignment with target defect semantics. Human Preference Score (HPS) is used to reflect subjective editing quality, while LPIPS measures perceptual differences in local regions, indicating editing strength. Notably, LPIPS is used in both comparative and ablation experiments with different objectives. In comparison, it measures overall similarity between the original and edited images, where lower values indicate better preservation of object identity. In ablation, it reflects the extent of local attribute editing, with higher values indicating more thorough removal and a corresponding category shift.}

\textcolor{rv1}{At present, there is no standardized and comprehensive evaluation metric in the image editing domain. Human evaluation has been widely adopted as an important reference, and many existing studies use consistency with human judgment as a standard metric \cite{ref-table}. Therefore, introducing subjective evaluation in this work is justified. Unlike attributes such as age that are difficult to define precisely, bolt defects are more standardized and perceptually distinct, making subjective evaluation more applicable and reliable in this task. To this end, we designed HPS to better align with human visual preferences.} The specific rules are as follows: for defect images edited by \( M \) sets of ablation experiment configurations, an expert group ranks them and assigns scores \( S_{i,k} = \{1, 2, \dots, M\} \), where each score is unique. A higher score indicates better editing performance. HPS is defined \textcolor{fn}{as follows}:
\begin{equation}
\begin{aligned}
\mathrm{HPS}=  \frac{ {\textstyle \sum_{i=1}^{N} \sum_{k=1}^{K} S_{i,k}}}{N \cdot K}
\end{aligned}
\label{eq11}
\end{equation}
\noindent where \( S_{i,k} \) represents the score given by expert \( i \) to the \( k \)-th image, \( N \) denotes the number of experts, \( K \) represents the total number of edited images, and \( M \) refers to the number of ablation experiment configurations.

\subsection{Ablation Study}
To verify the specific contributions of each module in our method, we conducted two ablation studies, evaluating the segmentation performance of Bolt-SAM and the editing effectiveness of SBDE. All experiments were performed on the BAS dataset.

\subsubsection{Ablation Study on Bolt-SAM Segmentation}
\begin{figure}[!t]
\centering
\includegraphics[width=0.8\linewidth]{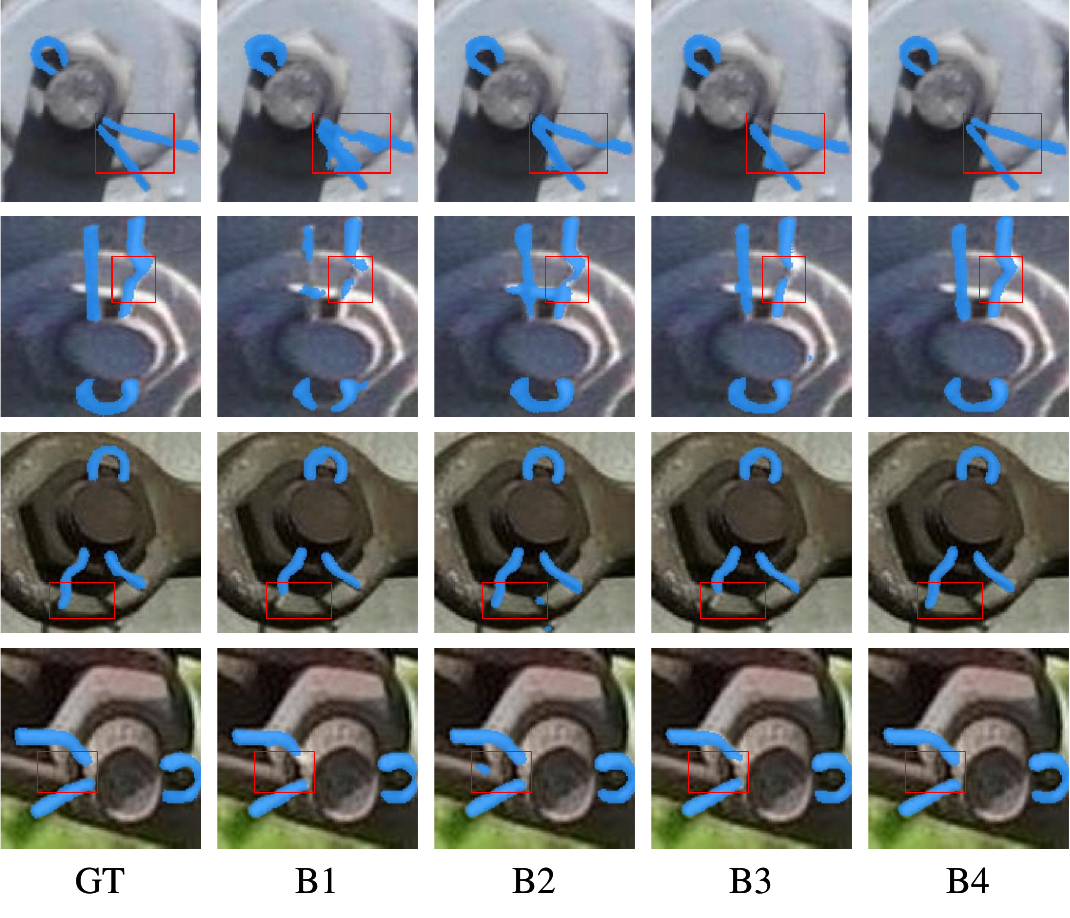}
\caption{Visualization of Ablation Results for Bolt-SAM Segmentation Based on Pin Attributes}
\label{fig6}
\vspace{-0.3cm}
\end{figure}

Bolt-SAM is the core algorithm of SBDE, focusing on generating high-quality bolt attribute masks. To validate the effectiveness of the CFA module and the MAMD architecture, we conducted segmentation ablation experiments, with quantitative results presented in Table \ref{tab4}. Compared with the baseline model (B1), the proposed modules showed significant improvements across all metrics. The CFA module (B2) demonstrated particularly notable enhancements in nut attributes, with mIoU, Dice, and PA increasing by 6.34, 4.57, and 1.95, respectively. This is because nuts are located at the bottom of bolts and are prone to shadowing effects caused by lighting, leading to blurred and indistinguishable edges, \textcolor{fn}{which are} effectively mitigated by CFA. MAMD (B3) is a multi-stage architecture designed for pins, which achieved more precise localization of pin components and outperformed B2 in terms of the PA metric. It is important to note that PA measures the overall pixel accuracy, and due to the small size of bolt images, the improvement in PA is relatively limited, especially for pin attributes. Overall, Bolt-SAM (B4) combines CFA and MAMD, balances the unique characteristics of bolt attributes, and achieves the best results across all metrics. Fig. \ref{fig6} shows the visualization results of the improved module on pin segmentation, among which B4 achieves the best segmentation effect.

\begin{table}[!t]
\caption{Results of Ablation Study on Bolt-SAM Segmentation. The Best Results are Marked in Bold.}
\label{tab4}
\centering
\begin{tabularx}{\linewidth}{>{\centering\arraybackslash}m{0.07\linewidth}>{\centering\arraybackslash}m{0.05\linewidth}>{\centering\arraybackslash}m{0.07\linewidth}>{\centering\arraybackslash}X>{\centering\arraybackslash}X>{\centering\arraybackslash}X>{\centering\arraybackslash}X>{\centering\arraybackslash}X>{\centering\arraybackslash}X}
\toprule
\multirow{2}{*}{Method} & \multirow{2}{*}{CFA} & \multirow{2}{*}{MAMD} & \multicolumn{3}{c}{\text {Pin} {(\%)}} & \multicolumn{3}{c}{\text {Nut} {(\%)}}  \\ \cmidrule(r){4-6} \cmidrule(r){7-9} 
                        &                      &                       & $\text {mIoU} \uparrow$   & $\text {Dice}  \uparrow$   & $\text {PA} \uparrow$     & $\text {mIoU} \uparrow$   & $\text {Dice} \uparrow$   & $\text {PA} \uparrow$     \\ \midrule
B1                      &                     &                      & 71.16 & 82.80  & 97.48 & 70.44 & 81.90  & 92.76 \\
B2                      & $\surd$                    &                      & 74.83 & 85.43 & 97.86 & \textbf{76.78} & \textbf{86.47} & \textbf{94.71} \\
B3                      &                     & $\surd$                     & 73.75 & 84.72 & 97.90  & -      & -      & -      \\
B4                      & $\surd$                    & $\surd$                     & \textbf{75.66} & \textbf{86.01} & \textbf{98.02} & -      & -      & -      \\ \bottomrule
\end{tabularx}
\end{table}

\begin{figure}[!t]
\centering
\includegraphics[width=0.7\linewidth]{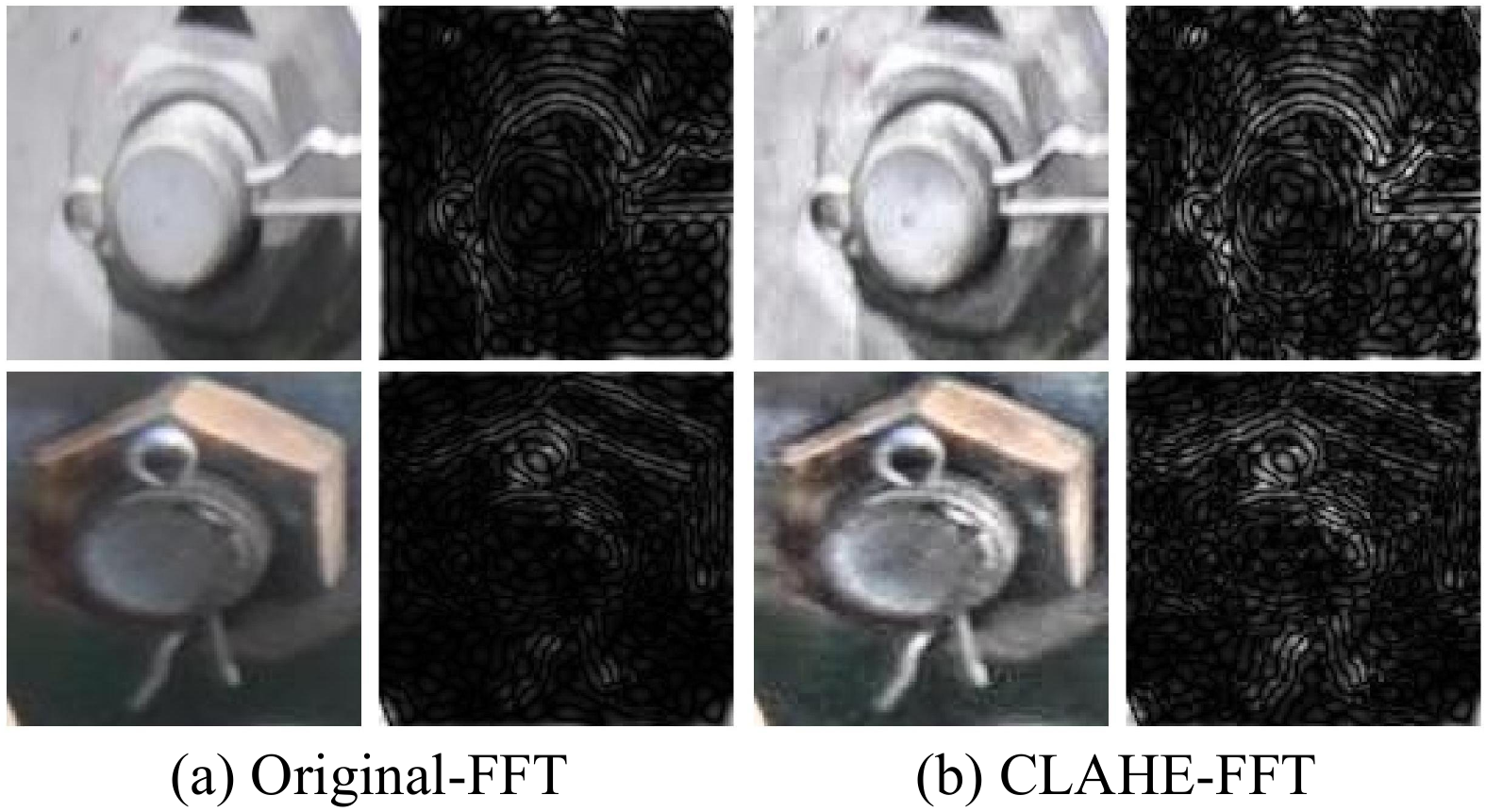}
\caption{\textcolor{rv1}{Visualization of the enhancement results by the CLAHE-FFT module. (a) Original image and its high-frequency components; (b) CLAHE-enhanced image and its high-frequency components.}}
\label{fig7}
\vspace{-0.3cm}
\end{figure}

To show the effectiveness of the CFA module in enhancing bolt edge details, we present the visualization results before and after CLAHE processing, along with their corresponding high-frequency components, in Fig. \ref{fig7}. The results indicate that CLAHE significantly enhances structural details in low-contrast regions, particularly when the bolt and the background \textcolor{fn}{fittings} have similar colors. In such cases, the edges of the nut and the pin regions appear noticeably clearer and more complete in the high-frequency components. \textcolor{rv2}{In addition, we also compare different adapter configurations under the same experimental conditions, as shown in Table \ref{tab-cfa}. The results show that CFA achieves the best performance in terms of mIoU, Dice, and PA for both the Pin and Nut attributes, demonstrating a more stable and significant improvement in overall segmentation accuracy compared with other adapters. This advantage mainly benefits from CLAHE’s adaptive enhancement of lighting and contrast before feature extraction, which effectively alleviates edge blurring.}

\begin{table}[!t]
{\color{rv2}
\caption{Comparison of different adapter modules in Bolt-SAM.}
\label{tab-cfa}
\centering
\begin{tabularx}{\linewidth}{>{\centering\arraybackslash}m{0.23\linewidth}>{\centering\arraybackslash}X>{\centering\arraybackslash}X>{\centering\arraybackslash}X>{\centering\arraybackslash}X>{\centering\arraybackslash}X>{\centering\arraybackslash}X}
\toprule
\multirow{2}{*}{Adapter} & \multicolumn{3}{c}{Pin {(\%)}}                            & \multicolumn{3}{c}{Nut {(\%)}}                             \\ \cmidrule(r){2-4} 
\cmidrule(r){5-7}
                        & $\text {mIoU} \uparrow$   & $\text {Dice} \uparrow$   & $\text {PA} \uparrow$              & $\text {mIoU} \uparrow$   & $\text {Dice} \uparrow$   & $\text {PA} \uparrow$              \\ \midrule
None              & 73.75          & 84.72          & 97.90          & 70.44          & 81.90          & 92.76          \\
Laplacian \cite{ref-evp}        & 74.20          & 85.10          & 97.93          & 72.15          & 83.15          & 93.40          \\
FFT \cite{ref-samadapter}              & 75.05          & 85.78          & 97.97          & 74.80          & 85.35          & 94.15          \\
CLAHE-FFT         & \textbf{75.66} & \textbf{86.01} & \textbf{98.02} & \textbf{76.78} & \textbf{86.47} & \textbf{94.71} \\ \bottomrule
\end{tabularx}
}
\end{table}

\subsubsection{Ablation Study on SBDE Editing}
SBDE achieves precise editing of bolt defect attributes by combining Bolt-SAM and MOD-LaMa. \textcolor{rv1}{While Bolt-SAM has been previously analyzed, we here assess the reliability of structural element settings in MOD-LaMa. We compare the editing performance using different structural element sizes under an automated pipeline and introduce a manual adjustment strategy as a reference. This strategy filters out failed automated samples and re-edits them by manually adjusting the structuring elements until satisfactory results are obtained. As shown in Table \ref{tab5}, the default setting used in this study (3 for pins and 5 for nuts) achieves CSFIDs that are already very close to those of manual adjustment, and performs even better in LPIPS. Manual adjustment typically expands the structuring element to cover broader regions, resulting in increased LPIPS. These results confirm that MOD-LaMa ensures high editing quality under a unified setting while supporting efficient batch processing, validating its reliability within the SBDE framework.}
\begin{table}[!t]
{\color{rv1}
\caption{Comparison of editing results under automated processing with different structuring elements and manual adjustment.}
\label{tab5}
\begin{tabularx}{\linewidth}{>{\centering\arraybackslash}X>{\centering\arraybackslash}X>{\centering\arraybackslash}X>{\centering\arraybackslash}X>{\centering\arraybackslash}X>{\centering\arraybackslash}X}
\toprule
\multirow{2}{*}{\makecell{Structuring \\Element}} & \multirow{2}{*}{Operation} & \multicolumn{2}{c}{Pin}         & \multicolumn{2}{c}{Nut}         \\ \cmidrule(r){3-4} \cmidrule(r){5-6} 
                                          &                            & CSFID↓         & LPIPS↓{(\%)}         & CSFID↓         & LPIPS↓{(\%)}         \\ \midrule
0                                         & \multirow{5}{*}{Automatic} & 17.32          & \textbf{15.43} & 19.95          & \textbf{10.90} \\
1                                         &                            & 18.34          & 16.80          & 21.26          & 11.06          \\
3                                         &                            & 16.10          & 19.05          & 19.12          & 12.82          \\
5                                         &                            & 16.37          & 21.64          & 18.95          & 14.97          \\
7                                         &                            & 16.76          & 23.06          & 19.37          & 16.43          \\ \midrule
Multiple                                     & Manual                     & \textbf{16.05} & 20.47          & \textbf{18.87} & 15.49          \\ \bottomrule
\end{tabularx}
}
\vspace{-0.3cm}
\end{table}

\begin{figure}[!t]
\centering
\includegraphics[width=0.8\linewidth]{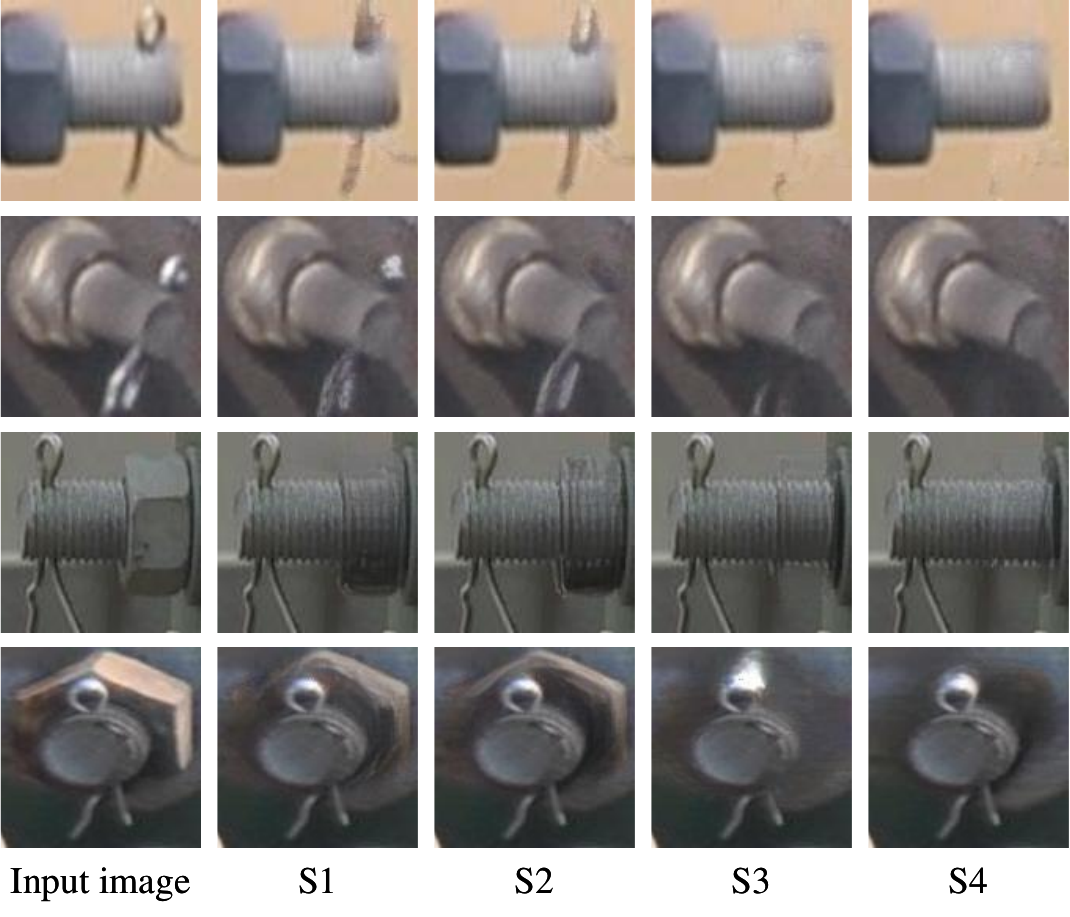}
\caption{Visualization of Ablation Results for SBDE Editing}
\label{fig8}
\vspace{-0.3cm}
\end{figure}

To validate the contributions of these two core modules, we designed ablation experiments for the editing task, using \textcolor{rv1}{CSFID, HPS, and LPIPS} as evaluation metrics, as described in \textcolor{rv1}{Section \ref{sc-metric}}. The experiments included four ablation configurations, with each configuration generating 300 edited images. The scores were provided by 20 domain experts. Quantitative results are presented in Table \ref{tab6}, and Fig. \ref{fig8} shows a visual comparison of editing results across different module configurations.

\begin{table}[!t]
\caption{\textcolor{rv1}{Results of Ablation Study on SBDE Editing. LPIPS is reported as $\times 10^2$.}}
\label{tab6}
\centering
\begin{tabularx}{\linewidth}{>{\centering\arraybackslash}X>{\centering\arraybackslash}X>{\centering\arraybackslash}X>{\centering\arraybackslash}X>{\centering\arraybackslash}X>{\centering\arraybackslash}X>{\centering\arraybackslash}X>{\centering\arraybackslash}X>{\centering\arraybackslash}X}
\toprule
\multirow{2}{*}{Method} & \multirow{2}{*}{\makecell{Bolt-\\SAM}} & \multirow{2}{*}{\makecell{MOD-\\LaMa}} & \multicolumn{3}{c}{Pin}                         & \multicolumn{3}{c}{Nut}                         \\ \cmidrule(r){4-6} \cmidrule(r){7-9} 
                        &                      &                      & \textcolor{rv1}{CSFID↓}         & HPS↑          & \textcolor{rv1}{LPIPS↑}         & \textcolor{rv1}{CSFID↓}         & HPS↑          & \textcolor{rv1}{LPIPS↑}         \\ \midrule
S1                      &                      &                      & \textcolor{rv1}{17.49}          & 1.17          & \textcolor{rv1}{11.84}          & \textcolor{rv1}{19.37}          & 1.08          & \textcolor{rv1}{10.53}          \\
S2                      & $\surd$              &                      & \textcolor{rv1}{18.02}          & 1.83          & \textcolor{rv1}{11.59}          & \textcolor{rv1}{19.62}          & 2.14          & \textcolor{rv1}{10.42}          \\
S3                      &                      & $\surd$              & \textcolor{rv1}{17.27}          & 3.20          & \textcolor{rv1}{\textbf{19.54}} & \textcolor{rv1}{19.12}          & 2.87          & \textcolor{rv1}{\textbf{15.38}} \\
S4                      & $\surd$              & $\surd$              & \textcolor{rv1}{\textbf{16.10}} & \textbf{3.80} & \textcolor{rv1}{19.05}          & \textcolor{rv1}{\textbf{18.95}} & \textbf{3.91} & \textcolor{rv1}{14.97}          \\ \bottomrule
\end{tabularx}
\end{table}

\textcolor{rv1}{The baseline (S1), which combines RobustSAM and LaMa without any enhancements, yields the worst performance, with the lowest HPS and LPIPS, indicating limited attribute changes and indistinct defect features. Its relatively high CSFID further reflects poor semantic alignment with the target category. Introducing Bolt-SAM (S2) improves segmentation and slightly raises HPS, but LPIPS remains similar and CSFID increases, suggesting that more precise masks do not yet ensure effective attribute removal. With the addition of MOD-LaMa (S3), editing performance improves significantly. HPS and LPIPS rise markedly, and CSFID decreases, indicating that morphological optimization plays a key role in enhancing attribute removal. Finally, SBDE (S4), integrating both modules, achieves the best overall performance. Specifically, CSFID decreases to 16.10 and 18.95, LPIPS improves to 19.05 and 14.97, and HPS reaches 3.80 and 3.91 for pin and nut defects, respectively. Visual results in Fig. \ref{fig8} further confirm these findings.}

\textcolor{rv1}{In summary, Bolt-SAM enhances mask precision, improving localization of the editing region, while MOD-LaMa optimizes the mask and strengthens attribute removal. The synergy of both modules markedly enhances SBDE's performance in bolt defect attribute editing tasks.}

\subsection{Comparison with Existing Methods}
To comprehensively validate the performance advantages of the SBDE in segmentation and editing, we conducted detailed comparisons with state-of-the-art segmentation and editing methods. The comparative experiments are divided into two parts: segmentation performance comparison for Bolt-SAM and editing effectiveness comparison for SBDE.

\subsubsection{Comparative Experiment of Bolt-SAM Segmentation}
In the segmentation, we compared Bolt-SAM with SAM, SA, HS, and RS, while also exploring the impact of the number of point prompts on bolt attribute segmentation performance. It is important to note that point prompts refer to the number of positive prompt points set for each part of the pin. The experiments are conducted on the BAS dataset, with all comparison methods fine-tuned. Segmentation performance is evaluated using three metrics: mIoU, Dice, and PA.

The quantitative results, as shown in Table \ref{tab7}, indicate that Bolt-SAM significantly outperforms all other methods across all metrics, achieving the best performance when using the \textcolor{fn}{3-point} prompt (which is adopted as the default setting in all other experiments in this paper). Compared with the baseline, Bolt-SAM improves the mIoU, Dice, and PA in pin segmentation by 4.50, 3.21, and 0.54, respectively, and improves the nut segmentation by 6.34, 4.57, and 1.95.

\begin{table}[!t]
\caption{Results of Comparative Experiment on Bolt-SAM Segmentation. 1p and 3p indicate the segmentation results using 1-point and 3-point prompts, respectively.}
\label{tab7}
\centering
\begin{tabularx}{\linewidth}{>{\centering\arraybackslash}m{0.23\linewidth}>{\centering\arraybackslash}X>{\centering\arraybackslash}X>{\centering\arraybackslash}X>{\centering\arraybackslash}X>{\centering\arraybackslash}X>{\centering\arraybackslash}X}
\toprule
\multirow{2}{*}{Method} & \multicolumn{3}{c}{Pin {(\%)}}                            & \multicolumn{3}{c}{Nut {(\%)}}                             \\ \cmidrule(r){2-4} 
\cmidrule(r){5-7}
                        & $\text {mIoU} \uparrow$   & $\text {Dice} \uparrow$   & $\text {PA} \uparrow$              & $\text {mIoU} \uparrow$   & $\text {Dice} \uparrow$   & $\text {PA} \uparrow$              \\ \midrule
SAM-1p \cite{ref-sam}            &  72.06          &  82.82          &  97.21          &  75.17          &  85.29          &  94.09          \\
SAM-3p \cite{ref-sam}           &  71.42          &  82.35          &  97.00            &  75.34          &  85.40           &  94.12          \\
SA-1p \cite{ref-samadapter}    &  62.37          &  75.75          &  96.93          &  41.56          &  53.49          &  87.47          \\
SA-3p \cite{ref-samadapter}   &  67.79          &  80.08          &  97.20           &  51.30           &  63.86          &  88.59          \\
HS-1p \cite{ref-hqsam}         &  60.40           &  74.51          &  96.60           &  57.02          &  70.85          &  87.04          \\
HS-3p \cite{ref-hqsam}        &  65.28          &  78.33          &  96.68          &  59.10           &  72.71          &  87.05          \\
RS-1p \cite{ref-rbsam}      &  69.15          &  81.35          &  97.38          &  69.51          &  81.25          &  92.64          \\
RS-3p \cite{ref-rbsam}     &  71.16          &  82.80           &  97.48          &  70.44          &  81.90           &  92.76          \\
ours-1p        &  72.73          &  82.88          &  97.61          &  75.24          &  85.31          &  94.25          \\
ours-3p       & \textbf{75.66} & \textbf{86.01} & \textbf{98.02} & \textbf{76.78} & \textbf{86.47} & \textbf{94.71} \\ \bottomrule
\end{tabularx}
\vspace{-0.3cm}
\end{table}

Further analysis of the impact of the number of point prompts on model performance shows that increasing the number of prompts generally improves segmentation results. This effect is particularly evident in Bolt-SAM's pin segmentation, where using the \textcolor{fn}{3-point} prompt improves mIoU, Dice, and PA by 2.93, 3.13, and 0.41, respectively, compared to a 1-point prompt. This is because the pixels of the bolt pins are too small, and more prompts can help the model better capture the boundaries of the target region. This is the opposite of SAM's pin segmentation, reflecting SAM's strong generalization.

\begin{figure}[!t]
\centering
\includegraphics[width=1\linewidth]{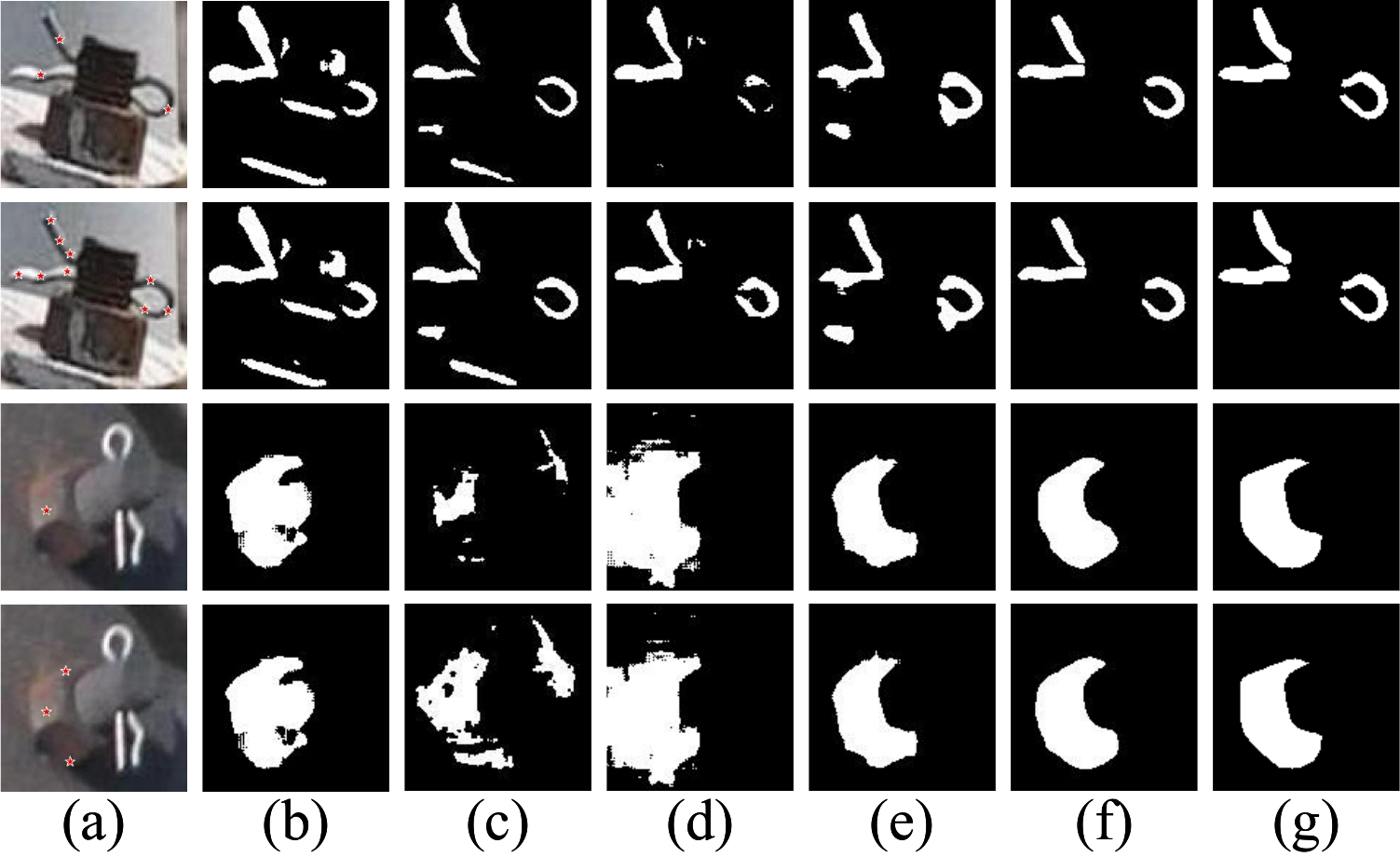}
\caption{Visualization of Comparative Experiment for Bolt-SAM Segmentation. \textcolor{fn}{(a) Input image; (b) SAM; (c) SA; (d) HS; (e) RS; (f) Ours; (g) Ground truth.} Red pentagrams indicate point prompts. Each group shows segmentation results using 1-point and 3-point prompts per region, respectively.}
\label{fig9}
\vspace{-0.3cm}
\end{figure}

Fig. \ref{fig9} shows the segmentation results of different methods. It can be observed that the masks generated by Bolt-SAM have clearer boundaries and are highly consistent with the ground truth masks. Even under complex background interference, Bolt-SAM accurately captures the edge information of bolt attributes, while other methods exhibit obvious boundary blurring and missed and false detection of target regions.

\subsubsection{Comparative Experiment of SBDE Editing}
The editing principle of SBDE is different from mainstream editing methods. Since no similar approaches currently exist, we compare SBDE with state-of-the-art \textcolor{rv2}{GAN-based, diffusion-based, and flow-based} editing methods, including SG, E4e, NT, NP, PnP, \textcolor{rv2}{FlowEdit, and RF-Edit. To ensure fairness, all methods are evaluated on the same BDG dataset using PSNR, SSIM, LPIPS, and KID as metrics. Each comparison method follows its officially released training or fine-tuning paradigm, and the editing experiments are conducted under a unified data split and preprocessing pipeline. In contrast, SBDE performs defect editing by removing bolt attributes through a cross-task joint framework that leverages a pre-trained image inpainting model. The entire process is accomplished solely based on normal samples, which represents a more rigorous experimental setting and better demonstrates the advantages of the proposed method.}

The quantitative results in Table \ref{tab8} demonstrate that SBDE outperforms the base models significantly across all evaluation metrics. In terms of image similarity, \textcolor{rv2}{compared with the current best-performing \textcolor{fn}{method, RF-Edit}, SBDE improves PSNR and SSIM by 8.10 and 3.63, respectively, and reduces LPIPS by 21.10 for pin defect editing. For nut defect editing, it increases PSNR and SSIM by 3.88 and 2.51, respectively, and reduces LPIPS by 19.29.} These results indicate that SBDE enables effective attribute editing while maintaining global structural consistency in the image. In terms of image quality and diversity, SBDE achieves KID scores of 56.19 and 44.92 for pin and nut defect editing, respectively, which are significantly lower than other methods. This suggests that the SBDE-edited images are closer in distribution to real samples, making them more suitable for downstream tasks.

\begin{figure*}[!t]
\centering
\includegraphics[width=0.8\linewidth]{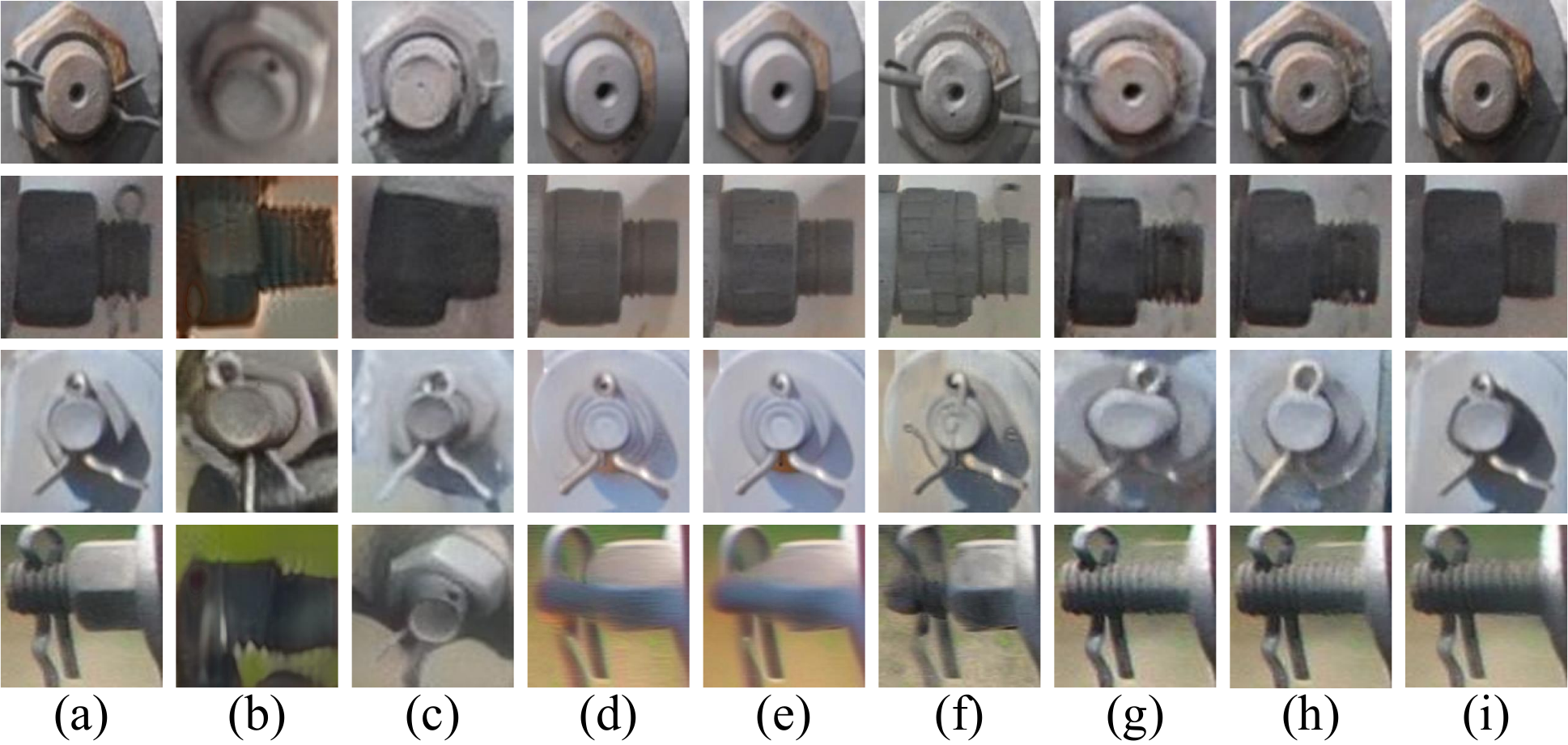}
\caption{\textcolor{rv2}{Visualization of Comparative Experiment for SBDE Editing. \textcolor{fn}{(a) Input image; (b) SG; (c) E4e; (d) NT; (e) NP; (f) PnP; (g) FlowEdit; (h) RF-Edit; (i) SBDE.} The top two rows show pin editing results, and the bottom two rows show nut editing results.}}
\label{fig10}
\vspace{-0.3cm}
\end{figure*}

Fig. \ref{fig10} shows a visual comparison of editing results across different methods. SBDE generates defect images that are highly consistent with the realistic perception of defects, with the natural target region editing and intact backgrounds. In contrast, SG and E4e tend to search for latent vectors in the generative space that semantically align with the corresponding defect, resulting in significant deviations from the original image and poor convergence. \textcolor{rv2}{Although NT, NP, PnP, FlowEdit, and RF-Edit maintain high fidelity across the entire image, their editing performance in the defect regions is unsatisfactory, often leading to residual attribute details or artifacts. This is mainly caused by domain discrepancies during the fine-tuning of Stable Diffusion and Flux, as well as the limitations introduced by differences between fine-tuning data and pre-training data, which restrict the applicability of large model-based methods in industrial defect editing tasks.}

In addition to editing quality, Table \ref{tab8} reports the inference time of each method to assess their suitability for industrial-scale applications. SBDE achieves the best editing performance while requiring significantly less inference time than \textcolor{rv2}{diffusion-based and flow-based} methods. Although GAN-based editing methods are faster, their editing results are unacceptable. Moreover, SBDE supports deployment on GPU and even CPU devices, demonstrating potential for industrial-scale deployment.

\begin{table*}[!t]
\caption{\textcolor{rv1}{Results of the Comparative Experiments on SBDE Editing, with Average Single-image Inference Time over 10 Runs on 128×128 Images using a Single NVIDIA RTX 4090 GPU.}}
\label{tab8}
\centering
\begin{tabularx}{\linewidth}{>{\centering\arraybackslash}m{0.1\linewidth}
>{\centering\arraybackslash}X >{\centering\arraybackslash}X 
>{\centering\arraybackslash}X >{\centering\arraybackslash}X 
>{\centering\arraybackslash}X >{\centering\arraybackslash}X 
>{\centering\arraybackslash}X >{\centering\arraybackslash}X>{\centering\arraybackslash}X>{\centering\arraybackslash}X}
\toprule
\multirow{2}{*}{Method} & \multirow{2}{*}{Generator} & \multirow{2}{*}{\textcolor{rv1}{Time(s)}} & \multicolumn{4}{c}{Pin}                                           & \multicolumn{4}{c}{Nut}                                           \\ \cmidrule(r){4-7} \cmidrule(r){8-11}
                        &                            &                          & SSIM↑{(\%)}          & PSNR↑          & LPIPS↓{(\%)}         & \textcolor{rv1}{KID↓{(\textperthousand)}}           & SSIM↑{(\%)}          & PSNR↑          & LPIPS↓{(\%)}         & \textcolor{rv1}{KID↓{(\textperthousand)}}           \\ \midrule
SG\cite{ref-stargan2}                      & -                          & \textcolor{rv1}{\textbf{0.11}}            & 40.52          & 11.58          & 89.22          & \textcolor{rv1}{249.79}         & 32.97          & 11.04          & 78.44          & \textcolor{rv1}{272.63}         \\
E4e\cite{ref-e4e}                     & Ada\cite{ref-ada}                        & \textcolor{rv1}{0.17}                     & 53.64          & 16.1\textcolor{fn}{0}           & 62.51          & \textcolor{rv1}{218.65}         & 49.63          & 15.92          & 56.76          & \textcolor{rv1}{257.93}         \\ \midrule
NT\cite{ref-null}                      & \multirow{3}{*}{\makecell{SD\cite{ref-sd}\\+DB\cite{ref-db}}}     & \textcolor{rv1}{37.46}                    & 81.46          & 17.72          & 46.03          & \textcolor{rv1}{128.10}         & 79.01          & 19.07          & 36.64          & \textcolor{rv1}{139.69}         \\
NP\cite{ref-nega}                      &                            & \textcolor{rv1}{5.12}                     & 80.31          & 17.39          & 48.24          & \textcolor{rv1}{116.03}         & 78.85          & 18.19          & 39.23          & \textcolor{rv1}{120.77}         \\
PnP\cite{ref-pnp}                     &                            & \textcolor{rv1}{8.37}                     & 79.13          & 17.19          & 50.55          & \textcolor{rv1}{131.06}         & 78.23          & 18.12          & 40.07          & \textcolor{rv1}{128.72}         \\ \midrule
\textcolor{rv2}{FlowEdit\cite{ref-flowedit}} &                       & \textcolor{rv2}{11.81} & \textcolor{rv2}{80.84} & \textcolor{rv2}{19.91} & \textcolor{rv2}{40.15} & \textcolor{rv2}{92.47} & \textcolor{rv2}{79.51} & \textcolor{rv2}{19.44} & \textcolor{rv2}{34.26} & \textcolor{rv2}{105.24} \\
\textcolor{rv2}{RF-Edit\cite{ref-rfedit}}  & \multirow{-2}{*}{\textcolor{rv2}{Flux\cite{ref-flux}}} & \textcolor{rv2}{16.30}  & \textcolor{rv2}{82.31} & \textcolor{rv2}{21.71} & \textcolor{rv2}{37.39} & \textcolor{rv2}{84.35} & \textcolor{rv2}{80.20}  & \textcolor{rv2}{20.19} & \textcolor{rv2}{31.53} & \textcolor{rv2}{97.58}  \\ \midrule
SBDE                    & -                          & \textcolor{rv1}{0.31}                     & \textbf{88.94} & \textbf{23.54} & \textbf{19.05} & \textcolor{rv1}{\textbf{56.19}} & \textbf{83.39} & \textbf{21.95} & \textbf{14.97} & \textcolor{rv1}{\textbf{44.92}} \\ \bottomrule
\end{tabularx}
\end{table*}

\begin{table}[!t]
\caption{Data Distribution of the Training Set after Augmentation with Different Methods.}
\label{tab9}
\begin{tabularx}{\linewidth}{>{\centering\arraybackslash}X>{\centering\arraybackslash}X>{\centering\arraybackslash}X>{\centering\arraybackslash}X>{\centering\arraybackslash}X>{\centering\arraybackslash}X}
\toprule
\multirow{2}{*}{Data} & \multirow{2}{*}{\makecell{Inspection \\  images}} & \multicolumn{4}{c}{Instance images}            \\ \cmidrule{3-6} 
                         &                                      & NR & PL & NL & All   \\ \midrule
Original                 & 1433                                 & 3961   & 449        & 244        & 4654  \\
Augment                 & +759                                 & +1393  & +1179      & +528       & +3100 \\ \bottomrule
\end{tabularx}
\vspace{-0.3cm}
\end{table}

\subsection{Bolt Defect Detection with ERA}
To further validate the effectiveness of SBDE and the practical application value of its edited defect images in downstream tasks, we propose the ERA strategy to augment the BDD dataset and evaluate performance on several representative detection and classification models. Specifically, we screened normal bolt instances with a resolution greater than $64 \times 64$ \textcolor{rv2}{and clear structural details} from the BDD dataset as editable objects. \textcolor{rv2}{At the same time, inspection images containing fewer bolts were preferentially chosen as editing carriers to ensure that the editable regions were visually clear and to reduce the number of normal bolts added along with the entire image. On this basis, a small number of bolts were randomly selected for editing, maintaining an approximate 2:1 ratio between PL and NL defects to match the real-world defect distribution.} After preprocessing, these instances were transformed into defective bolts using SBDE, and their corresponding labels were modified to create new inspection images with defects, as illustrated in Fig. \ref{fig1}. 

We evaluated the effectiveness of SBDE for data augmentation in downstream tasks. Using the ERA strategy, edited defect regions were restored into the original inspection images to generate augmented samples. \textcolor{rv2}{In practice, augmentation was performed on a full inspection image basis: edited bolts were counted under their corresponding defect categories, while unedited normal bolts in the same image were included in NR. The distribution of augmented results following this process is shown in Table \ref{tab9}. To avoid data leakage, all editing and augmentation were conducted only on the training set, and the test set remained completely independent and unmodified. In addition, all methods adopted moderate augmentation with a consistent number of samples to prevent redundant data and distribution shift caused by excessive augmentation \cite{ref-data}.}

To ensure comprehensive evaluation, we conducted experiments across three representative model architectures, covering both detection and classification tasks: 1) Vision Transformer models, including ViTDet \cite{ref-vitdet} for detection and ViT \cite{ref-vit} for classification; 2) attention-based CNNs, including YOLOv11 \cite{ref-yolo11} for detection and ACmix \cite{ref-acmix} for classification; and 3) self-supervised models, using the pre-trained SimCLR \cite{ref-simclr} to construct Faster R-CNN \cite{ref-faster} for detection and ResNet-50 \cite{ref-res} for classification. It is important that these experiments aim to verify the impact of SBDE-augmented data on model performance, rather than optimizing any specific architecture. \textcolor{rv2}{To ensure fairness, all models used the default hyperparameter settings from their original papers under consistent experimental conditions. We focused on the relative improvements achieved by different augmentation methods within the same model, rather than comparing absolute performance differences across models. }As a data-level strategy, SBDE is compatible with structure-level methods, and the two can be jointly applied to further enhance detection and classification performance.

\textcolor{rv1}{Table \ref{tab10} shows the quantitative evaluation results of different detection models. SBDE-based augmentation consistently achieved the best performance, with mAP50 improvements of 3.4, 4.4, and 4.4 over ViTDet, YOLOv11, and Faster R-CNN, respectively, significant gains on the detection of defect categories. This improvement is largely attributed to its ability to balance category distribution through targeted defect augmentation. In contrast, other augmentation methods generally led to degraded performance, mainly due to the low quality of edited defect images and noticeable artifacts introduced during ERA-based restoration, which impaired box localization and affected detection.}

\textcolor{rv1}{Additionally, we conducted classification-based augmentation experiments, with the results summarized in Table \ref{tab11}. SBDE consistently achieved the best performance across all metrics, with IMP values reaching 7.2, 10.2, and 6.1 on ViT, Acmix, and SimCLR (ResNet-50), respectively, indicating significant improvements over the original baselines. In contrast, the performance of other methods varied considerably, with some even leading to performance degradation. This suggests that their edited images are of low quality, which weakens the semantic representation of defect categories and limits their effectiveness in improving classification performance.}

In summary, the proposed method \textcolor{rv2}{balances the training dataset} by editing and augmenting defect samples, \textcolor{rv2}{and under consistent and independent test conditions, it achieves performance improvements} across different tasks and models, \textcolor{rv2}{objectively verifying its} effectiveness, compatibility, and strong potential for real-world engineering applications.
\vspace{-0.3cm}
\textcolor{rv1}{\subsection{Generalization and Robustness Evaluation}}
\textcolor{rv1}{Due to the presence of varying environmental conditions and scene diversity in real-world applications, the robustness and generalization capability of the proposed method are critical for its practical effectiveness. We conduct experiments from two aspects: environmental adaptability and cross-domain transferability, to comprehensively evaluate its performance across diverse scenarios.}

\begin{table*}[!t]
{\color{rv1}
\caption{Comparison of mAP50 Detection Results across Different Augmentations.}
\label{tab10}
\begin{tabularx}{\textwidth}{>{\centering\arraybackslash}m{0.1\linewidth}>{\centering\arraybackslash}X>{\centering\arraybackslash}X>{\centering\arraybackslash}X>{\centering\arraybackslash}X>{\centering\arraybackslash}X>{\centering\arraybackslash}X>{\centering\arraybackslash}X>{\centering\arraybackslash}X>{\centering\arraybackslash}X>{\centering\arraybackslash}X>{\centering\arraybackslash}X>{\centering\arraybackslash}X}
\toprule
\multirow{2}{*}{Method} & \multicolumn{4}{c}{ViTDet} & \multicolumn{4}{c}{YOLOv11} & \multicolumn{4}{c}{SimCLR(Faster R-CNN)} \\ \cmidrule(r){2-5} \cmidrule(r){6-9} \cmidrule(r){10-13}
                        & NR    & PL   & NL   & mAP  & NR    & PL    & NL   & mAP  & NR       & PL       & NL       & mAP     \\ \midrule
Original                & 90.2  & 80.7 & 85.5 & 85.5 & 91.0  & 80.6  & 77.8 & 83.1 & 82.2     & 80.2     & 68.3     & 76.9    \\
+SG \cite{ref-stargan2}                    & 89.1  & 75.5 & 78.4 & 81.0 & 89.5  & 72.3  & 66.1 & 76.0 & 80.1     & 72.8     & 61.2     & 71.4    \\
+E4e \cite{ref-e4e}                    & 89.3  & 73.9 & 77.0 & 80.1 & 90.0  & 70.4  & 65.5 & 75.3 & 80.6     & 70.3     & 60.5     & 70.5    \\
+NT \cite{ref-null}                    & 90.1  & 78.0 & 83.3 & 83.8 & 90.8  & 78.5  & 74.2 & 81.2 & 82.8     & 76.1     & 71.4     & 76.8    \\
+NP \cite{ref-nega}                    & 90.4  & 79.1 & 84.7 & 84.7 & 91.1  & 79.7  & 75.1 & 82.0 & 83.1     & 76.7     & 72.3     & 77.4    \\
+PnP \cite{ref-pnp}                   & 89.9  & 77.2 & 82.4 & 83.2 & 90.5  & 78.2  & 73.0 & 80.6 & 81.6     & 75.4     & 64.2     & 73.7    \\
\textcolor{rv2}{+FlowEdit\cite{ref-flowedit}} & \textcolor{rv2}{90.5} & \textcolor{rv2}{83.5} & \textcolor{rv2}{86.3} & \textcolor{rv2}{86.8} & \textcolor{rv2}{90.7} & \textcolor{rv2}{83.9} & \textcolor{rv2}{77.2} & \textcolor{rv2}{83.9} & \textcolor{rv2}{83.1} & \textcolor{rv2}{78.9} & \textcolor{rv2}{70.1} & \textcolor{rv2}{77.4} \\
\textcolor{rv2}{+RF-Edit\cite{ref-rfedit}} & \textcolor{rv2}{90.8} & \textcolor{rv2}{84.4} & \textcolor{rv2}{87.0} & \textcolor{rv2}{87.4} & \textcolor{rv2}{90.9} & \textcolor{rv2}{85.1} & \textcolor{rv2}{78.6} & \textcolor{rv2}{84.9} & \textcolor{rv2}{83.5} & \textcolor{rv2}{79.6} & \textcolor{rv2}{71.2} & \textcolor{rv2}{78.1} \\
+SBDE                   & \textbf{91.7} & \textbf{85.9} & \textbf{89.1} & \textbf{88.9} & \textbf{91.2} & \textbf{89.3} & \textbf{81.9} & \textbf{87.5} & \textbf{86.5} & \textbf{83.6} & \textbf{73.9} & \textbf{81.3} \\ \bottomrule
\end{tabularx}}
\end{table*}

\begin{table*}[!t]
{\color{rv1}
\caption{Comparison of Classification Performance Metrics after Data Augmentation using Different Methods.}
\label{tab11}
\begin{tabularx}{\textwidth}{>{\centering\arraybackslash}m{0.1\linewidth}>{\centering\arraybackslash}X>{\centering\arraybackslash}X>{\centering\arraybackslash}X>{\centering\arraybackslash}X>{\centering\arraybackslash}X>{\centering\arraybackslash}X>{\centering\arraybackslash}X>{\centering\arraybackslash}X>{\centering\arraybackslash}X>{\centering\arraybackslash}X>{\centering\arraybackslash}X>{\centering\arraybackslash}X}
\toprule
\multirow{2}{*}{Method} & \multicolumn{4}{c}{ViT}                                      & \multicolumn{4}{c}{Acmix}                                     & \multicolumn{4}{c}{SimCLR(ResNet50)}                         \\ \cmidrule(r){2-5} \cmidrule(r){6-9} \cmidrule(r){10-13} 
                        & Acc           & macF1           & weightF1           & IMP          & Acc           & macF1           & weightF1           & IMP           & Acc           & macF1           & weightF1           & IMP          \\ \midrule
Original                & 75.3          & 59.5          & 71.2          & -            & 71.2          & 63.2          & 72.0          & -             & 78.2          & 66.3          & 75.9          & -            \\
+SG \cite{ref-stargan2}                    & 72.6          & 56.9          & 69.8          & -3.6         & 68.9          & 47.0          & 68.5          & -3.2          & 75.3          & 60.7          & 73.6          & -3.7         \\
+E4e \cite{ref-e4e}                   & 74.1          & 56.1          & 73.4          & -1.5         & 70.3          & 57.4          & 68.2          & -1.3          & 76.2          & 65.7          & 74.7          & -2.6         \\
+NT \cite{ref-null}                    & 77.1          & 61.7          & 77.5          & 2.4          & 74.1          & 60.6          & 72.7          & 4.1           & 78.2          & 66.1          & 77.0          & 0            \\
+NP \cite{ref-nega}                    & 75.7          & 61.5          & 74.6          & 0.6          & 73.0          & 54.4          & 70.4          & 2.6           & 78.0          & 62.1          & 77.2          & -0.3         \\
+PnP \cite{ref-pnp}                   & 74.6          & 59.4          & 73.3          & -0.9         & 72.3          & 54.6          & 70.8          & 1.6           & 77.1          & 65.0          & 75.9          & -1.4         \\
\textcolor{rv2}{+FlowEdit\cite{ref-flowedit}} & \textcolor{rv2}{79.0} & \textcolor{rv2}{69.8} & \textcolor{rv2}{79.2} & \textcolor{rv2}{4.9} & \textcolor{rv2}{76.8} & \textcolor{rv2}{65.9} & \textcolor{rv2}{76.4} & \textcolor{rv2}{7.9} & \textcolor{rv2}{80.1} & \textcolor{rv2}{70.9} & \textcolor{rv2}{80.3} & \textcolor{rv2}{2.4} \\
\textcolor{rv2}{+RF-Edit\cite{ref-rfedit}} & \textcolor{rv2}{79.8} & \textcolor{rv2}{71.5} & \textcolor{rv2}{80.0} & \textcolor{rv2}{6.0} & \textcolor{rv2}{78.3} & \textcolor{rv2}{69.2} & \textcolor{rv2}{78.0} & \textcolor{rv2}{10.0} & \textcolor{rv2}{81.5} & \textcolor{rv2}{74.1} & \textcolor{rv2}{81.6} & \textcolor{rv2}{4.2} \\
+SBDE                   & \textbf{80.7} & \textbf{75.9} & \textbf{80.7} & \textbf{7.2} & \textbf{81.4} & \textbf{73.8} & \textbf{80.9} & \textbf{10.2} & \textbf{83.0} & \textbf{78.8} & \textbf{83.1} & \textbf{6.1} \\ \bottomrule
\end{tabularx}}
\end{table*}

\subsubsection{\textcolor{rv1}{Robustness under Environmental Conditions}}
\textcolor{rv1}{Although the constructed dataset is derived from the real-world scenarios, some special environments (e.g., lighting, fog) may not be fully reflected. To further verify the applicability of the method in complex environments, we simulate five typical conditions based on the original inspection images, including bright, dark, foggy, rainy, and snowy scenarios, and edit the bolt attribute accordingly.}

\begin{figure}[!t]
\centering
\includegraphics[width=0.8\linewidth]{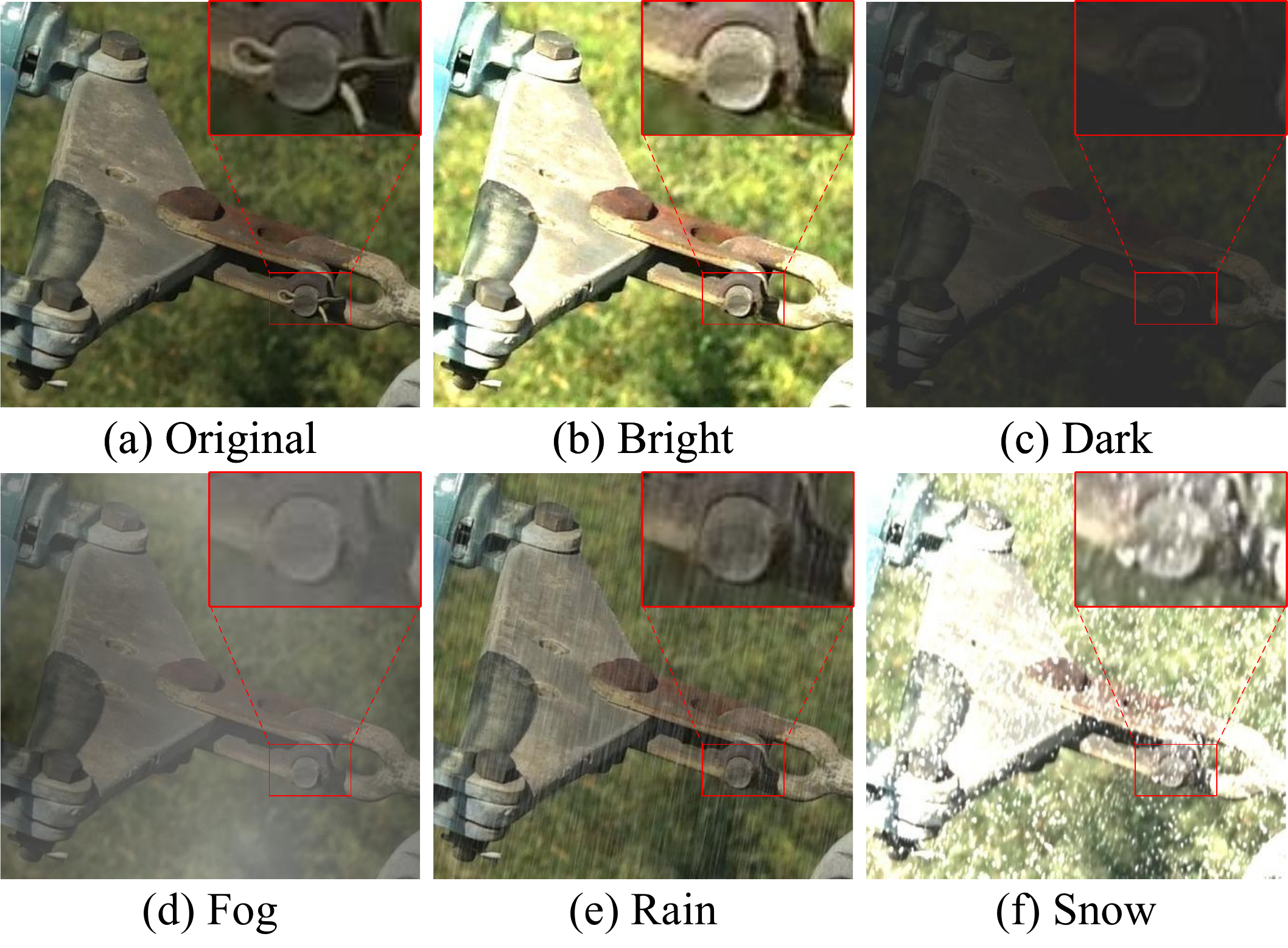}
\caption{\textcolor{rv1}{SBDE editing results under different environmental conditions. The original images are first subjected to environment simulation, followed by SBDE-based editing.}}
\label{fig11}
\vspace{-0.3cm}
\end{figure}

\textcolor{rv1}{Fig. \ref{fig11} illustrates the editing results of the proposed method under different environmental conditions. The results indicate high-quality edits across all simulated environments, with accurate bolt \textcolor{fn}{attribute} edits even under challenging conditions such as darkness or snow. In some cases, simulation-induced blurring enhances visual naturalness, indicating strong consistency and stability. As shown in Table \ref{tab12}, although metric fluctuations exist, performance remains within acceptable bounds. Notably, editing under dark and foggy conditions even surpasses the standard setting on certain metrics, further validating the method’s robustness across diverse scenarios.}

\begin{table}[!t]
{\color{rv1}
\caption{Quantitative Results of SBDE Editing under Different Environmental Conditions. SSIM and LPIPS are Reported as $\times 10^2$; KID is Reported as $\times 10^3$.}
\label{tab12}
\begin{tabularx}{\linewidth}{>{\centering\arraybackslash}X>{\centering\arraybackslash}X>{\centering\arraybackslash}X>{\centering\arraybackslash}X>{\centering\arraybackslash}X>{\centering\arraybackslash}X>{\centering\arraybackslash}X>{\centering\arraybackslash}X>{\centering\arraybackslash}X}
\toprule
\multirow{2}{*}{Method} & \multicolumn{4}{c}{Pin}                                           & \multicolumn{4}{c}{Nut}                                           \\ \cmidrule(r){2-5} \cmidrule(r){6-9}
                        & SSIM↑          & PSNR↑          & LPIPS↓         & KID↓           & SSIM↑          & PSNR↑          & LPIPS↓         & KID↓           \\ \midrule
\textcolor{rv2}{Original}                & 88.94          & 23.54          & 19.05          & 56.19          & 83.39          & 21.95          & 14.97          & 44.92          \\
Bright                  & 86.87          & 17.82          & 18.25          & 61.83          & 81.82          & 16.81          & 13.68          & 49.22          \\
Dark                    & \textbf{92.37} & \textbf{28.97} & 20.19          & \textbf{50.68} & \textbf{85.68} & \textbf{24.44} & 16.26          & \textbf{42.79} \\
Fog                     & 90.45          & 26.33          & 19.23          & 53.15          & 84.22          & 22.20          & 15.25          & 45.01          \\
Rain                    & 88.32          & 22.59          & 19.62          & 58.44          & 83.11          & 21.88          & 15.56          & 45.70          \\
Snow                    & 89.65          & 18.62          & \textbf{16.98} & 57.92          & 82.55          & 18.04          & \textbf{12.34} & 46.41          \\ \bottomrule
\end{tabularx}}
\vspace{-0.3cm}
\end{table}

\subsubsection{\textcolor{rv1}{Generalization on MVTec AD}}
To evaluate the reliability of the proposed method in defect editing tasks and its generalization capability beyond the transmission line domain, we conducted experiments on the publicly available industrial defect dataset MVTec AD. \textcolor{rv2}{We selected the transistor category with the cut lead defect and the cable category with the missing cable and missing wire defects as the targets. The former is characterized by missing or broken pins, which are highly similar to the bolt defects addressed in this study, while the latter correspond to the complete loss of a cable and the internal copper wire loss of a single cable, both of which are structural defects.}

\begin{figure*}[!t]
\centering
\includegraphics[width=0.9\linewidth]{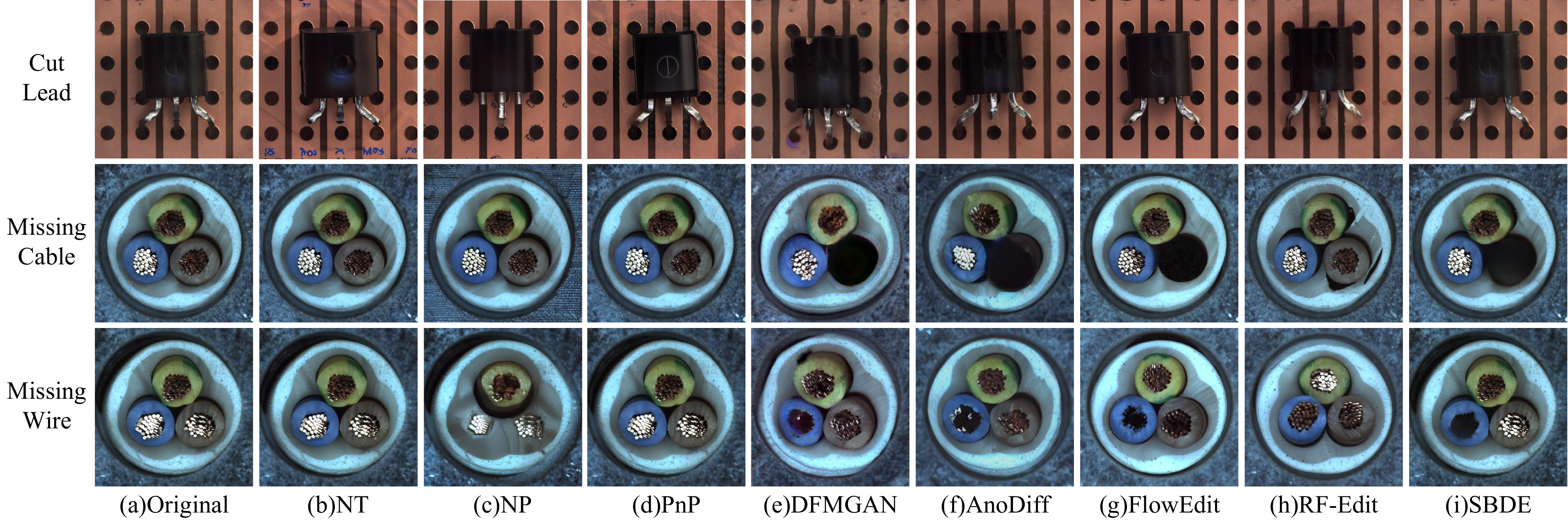}
\caption{\textcolor{rv2}{Visualization of defect editing results by different methods on the MVTec AD dataset.}}
\label{fig12}
\vspace{-0.3cm}
\end{figure*}

\begin{table}[!t]
{\color{rv2}
\caption{Quantitative results of different methods on the MVTec AD dataset, including three defect categories: Cut Lead, Missing Cable, and Missing Wire. KID is reported as $\times 10^3$ and LPIPS is reported as $\times 10^2$.}
\label{tab13}
\begin{tabularx}{\linewidth}{>{\centering\arraybackslash}m{0.19\linewidth}>{\centering\arraybackslash}X>{\centering\arraybackslash}X>{\centering\arraybackslash}X>{\centering\arraybackslash}X>{\centering\arraybackslash}X>{\centering\arraybackslash}X}
\toprule
                                & \multicolumn{2}{c}{Cut Lead}                                & \multicolumn{2}{c}{Missing Cable}                                             & \multicolumn{2}{c}{Missing Wire}                                              \\ \cmidrule{2-7} 
\multirow{-2}{*}{Method}        & KID↓                      & LPIPS↓                   & KID↓                               & LPIPS↓                            & KID↓                               & LPIPS↓                            \\ \midrule
NT\cite{ref-null}                              & 95.49                        & 29.22                        & \textcolor{rv2}{108.37}               & \textcolor{rv2}{33.08}                & \textcolor{rv2}{92.14}                & \textcolor{rv2}{30.25}                \\
NP\cite{ref-nega}                              & 98.13                        & 30.44                        & \textcolor{rv2}{112.26}               & \textcolor{rv2}{34.61}                & \textcolor{rv2}{125.03}               & \textcolor{rv2}{36.02}                \\
PnP\cite{ref-pnp}                             & 103.83                       & 27.04                        & \textcolor{rv2}{116.41}               & \textcolor{rv2}{29.58}                & \textcolor{rv2}{98.27}                & \textcolor{rv2}{28.12}                \\
DFMGAN\cite{ref-dfmgan}                          & 122.63                       & 32.32                        & \textcolor{rv2}{94.23}                & \textcolor{rv2}{35.52}                & \textcolor{rv2}{77.01}                & \textcolor{rv2}{34.48}                \\
AnoDiff\cite{ref-anodiff}                         & 73.31                        & 20.18                        & \textcolor{rv2}{79.12}                & \textcolor{rv2}{21.35}                & \textcolor{rv2}{70.84}                & \textcolor{rv2}{19.47}                \\
\textcolor{rv2}{FlowEdit\cite{ref-flowedit}}       & \textcolor{rv2}{58.75}       & \textcolor{rv2}{16.72}       & \textcolor{rv2}{\textbf{44.93}}       & \textcolor{rv2}{16.62}                & \textcolor{rv2}{40.91}                & \textcolor{rv2}{15.31}                \\
\textcolor{rv2}{RF-Edit\cite{ref-rfedit}}        & \textcolor{rv2}{75.46}       & \textcolor{rv2}{21.31}       & \textcolor{rv2}{82.17}                & \textcolor{rv2}{22.84}                & \textcolor{rv2}{86.03}                & \textcolor{rv2}{23.67}                \\
SBDE                            & \textbf{40.37}               & \textbf{15.85}               & \textcolor{rv2}{48.26}                & \textcolor{rv2}{\textbf{16.31}}       & \textcolor{rv2}{\textbf{37.58}}       & \textcolor{rv2}{\textbf{14.92}}       \\ \bottomrule
\end{tabularx}}
\vspace{-0.3cm}
\end{table}

We compared the proposed method with \textcolor{rv2}{several} representative approaches on this dataset, \textcolor{rv2}{including DFMGAN \cite{ref-dfmgan}, AnoDiff \cite{ref-anodiff}, FlowEdit \cite{ref-flowedit}, and RF-Edit \cite{ref-rfedit}}, while keeping all other experimental settings consistent. Fig. \ref{fig12} presents the editing results of different methods on the MVTec AD dataset. It can be observed that \textcolor{rv2}{in the cut lead defect, SBDE precisely removes the pins while maintaining overall structural consistency, clearly outperforming the other methods. In the missing cable defect, SBDE also achieves clear and natural editing results, although the regularity of the hole left after cable removal is slightly inferior to that of FlowEdit. In the missing wire task, SBDE performs better in fine-grained copper-core removal.} Furthermore, we followed the evaluation protocol adopted in related MVTec AD studies and conducted quantitative comparisons using KID and LPIPS. As shown in Table \ref{tab13}, SBDE achieved the best performance \textcolor{rv2}{in most categories}. The lowest KID score indicates superior overall image quality, while the minimal LPIPS value reflects more precise editing in the target region and minimal interference in non-target regions.

In summary, the proposed method demonstrates superior performance \textcolor{rv2}{on the standard public dataset, further validating its reliability, generalization, and reproducibility in cross-scenario applications. In addition, SBDE requires no modification to its core framework, and it can achieve precise mask localization across data domains simply by fine-tuning the segmentation module. Therefore, it maintains stable performance in different industrial scenarios and is} particularly suitable for industrial applications where defect samples are scarce. It holds potential applicability in structurally constrained industrial vision tasks, such as defect editing for bridge components and wind turbine parts.

\section{Conclusion}
\textcolor{md}{To alleviate the scarcity and imbalance of bolt defect samples in transmission lines, this paper proposes a segmentation-driven bolt defect editing method, SBDE, which converts normal bolts into defective ones for data augmentation and downstream tasks performance improvement. Specifically, Bolt-SAM, designed with the CFA and MAMD modules, enhances segmentation of complex bolt structures, while MOD-LaMa performs precise attribute removal for defect editing. The ERA strategy restores the edited bolts into inspection images to augment the defect dataset and improve defect detection accuracy. This approach provides an innovative solution to sample scarcity and shows potential for broader application.}

Although the proposed SBDE method has achieved good results in bolt editing, its performance remains influenced by the pixel quality of the original images. \textcolor{rv2}{In complex or low-quality images, noise and boundary blur may reduce local editing accuracy. Moreover, the method currently shows strong scene specificity and is mainly suitable for controllable one-part-off removal tasks, while broader editing scenarios may require incorporating additional priors.}

\textcolor{rv2}{In future work, we will integrate generative models to improve the quality and generalization of defect editing, and explore} the application of the method to other industrial vision tasks and end-to-end deployment to enhance its applicability in cross-domain scenarios. \textcolor{rv2}{We will also investigate segmentation and feature fusion strategies in transmission line scenes to support complex structure recognition and defect localization.}

\bibliographystyle{IEEEtran}
\bibliography{bibtex}

\vfill

\end{document}